\documentclass[lettersize,journal]{IEEEtran}

\usepackage{cite}
\usepackage{enumerate}
\usepackage{graphicx}
\usepackage{epsfig}
\usepackage{color}
\usepackage{threeparttable}
\usepackage{epstopdf}
\usepackage{multirow}
\usepackage{float}
\usepackage{bm}
\usepackage{verbatim}
\usepackage{makecell}
\usepackage{amsmath}

\usepackage{times}
\usepackage{array}
\usepackage{amssymb}
\usepackage{soul}
\usepackage{booktabs}
\usepackage{setspace}
\usepackage{tabularx}
\usepackage{diagbox} 
\usepackage{textcomp}
\usepackage{stfloats}
\usepackage{url}
\usepackage{amsfonts}
\usepackage{hyperref}
\usepackage{xcolor}
\usepackage{colortbl}
\usepackage{pifont}
\usepackage{subfigure}

\def\BibTeX{{\rm B\kern-.05em{\sc i\kern-.025em b}\kern-.08em
    T\kern-.1667em\lower.7ex\hbox{E}\kern-.125emX}}

\begin{document}

\title{OBSeg: Accurate and Fast Instance Segmentation Framework Using Segmentation Foundation Models with Oriented Bounding Box Prompts}

\author{Zhen Zhou$^{1, 2}$, 
Junfeng Fan$^{1}$,~\IEEEmembership{Senior Member,~IEEE,}
Yunkai Ma$^{1, 2, *}$,~\IEEEmembership{Member,~IEEE,}
Sihan Zhao$^{1, 2}$, 
Fengshui Jing$^{1, 2}$,~\IEEEmembership{Member,~IEEE,}
and Min Tan$^{1, 2}$
\thanks{This work was supported by the National Key R\&D Program of China (2023YFB4706800), the National Natural Science Foundation of China (62173327, 62373354), the Beijing Natural Science Foundation (4232057), and the Youth Innovation Promotion Association of CAS (2022130).}%
\thanks{$^{1}$The authors are with the State Key Laboratory of Multimodal Artificial Intelligence Systems, Institute of Automation, Chinese Academy of Sciences, Beijing, 100190, China. $^{2}$The authors are with the School of Artificial Intelligence, University of Chinese Academy of Sciences, Beijing, 100049, China. * Corresponding author. E-mail: yunkai.ma@ia.ac.cn}%
}

\markboth{}%
{OBSeg: Accurate and Fast Instance Segmentation Framework Using Segmentation Foundation Models with Oriented Bounding Box Prompts}

\maketitle

\begin{abstract}    
Instance segmentation in remote sensing images is a long-standing challenge. Since horizontal bounding boxes introduce many interference objects, oriented bounding boxes (OBBs) are usually used for instance identification. However, based on ``segmentation within bounding box'' paradigm, current instance segmentation methods using OBBs are overly dependent on bounding box detection performance. To tackle this problem, this paper proposes OBSeg, an accurate and fast instance segmentation framework using OBBs. OBSeg is based on box prompt-based segmentation foundation models (BSMs), e.g., Segment Anything Model. Specifically, OBSeg first detects OBBs to distinguish instances and provide coarse localization information. Then, it predicts OBB prompt-related masks for fine segmentation. Since OBBs only serve as prompts, OBSeg alleviates the over-dependence on bounding box detection performance of current instance segmentation methods using OBBs. Thanks to OBB prompts, OBSeg outperforms other current BSM-based methods using HBBs. In addition, to enable BSMs to handle OBB prompts, we propose a novel OBB prompt encoder. To make OBSeg more lightweight and further improve the performance of lightweight distilled BSMs, a Gaussian smoothing-based knowledge distillation method is introduced. Experiments demonstrate that OBSeg outperforms current instance segmentation methods on multiple datasets in terms of instance segmentation accuracy and has competitive inference speed. The code is available at \href{https://github.com/zhen6618/OBBInstanceSegmentation}{https://github.com/zhen6618/OBBInstanceSegmentation}.

\end{abstract}


\section{INTRODUCTION}
\IEEEPARstart {I}{nstance} segmentation of remote sensing images provides pixel-level foundational information for numerous tasks, such as maritime security \cite{Jstars_Orientated_Silhouette_Matching, Jstars_SAR_Ship_Instance}, agricultural measurement \cite{Jstars_An_Edge_Aware}, urban management, and disaster assessment. Pixel-level perception enables related instance segmentation methods to depict the exterior appearance and contour of the object. However, objects in remote sensing images usually have the characteristics of multi-scale, dense arrangement, and arbitrary orientations, which brings a long-standing challenge to the instance segmentation task.

\begin{figure}[t]
    \centering
    \centerline{\includegraphics[scale=0.666]{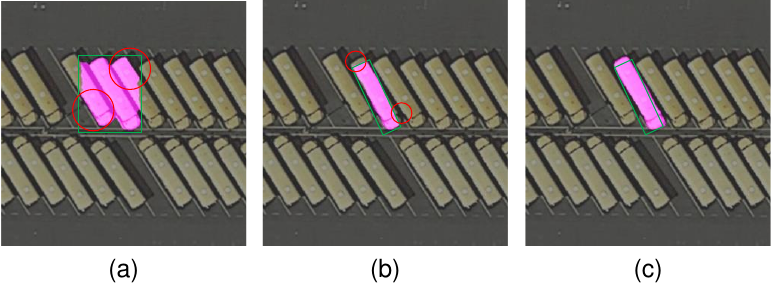}} 
    \caption{For instance segmentation in remote sensing images, (a): HBB introduces many interference objects. (b): The ``segmentation within bounding box'' paradigm limits the segmentation to be performed mainly within the detected OBB, making the segmentation performance overly dependent on the OBB detection performance. Once the OBB detection is inaccurate, the mask segmentation will also be affected. (c) The proposed OBSeg only uses OBB as a prompt to guide object segmentation, so the segmentation result is less dependent on OBB detection performance. Although the OBB detection is inaccurate, the mask can be segmented accurately.}
    \label{fig:Introduction}
    \vspace{-0.2cm}
\end{figure}

Since objects are densely packed in multiple orientations, which introduces many interference regions into horizontal bounding boxes (HBBs) \cite{HBB_Det_TIM, HBB_Det_TIM_2}, oriented bounding boxes (OBBs) are usually used for instance identification \cite{Rotated_Blend, OIS, Towards_Robust_Part, ISOP} (see Fig. \ref{fig:Introduction}). Existing instance segmentation methods are mainly based on the ``segmentation within bounding box'' paradigm, that is, object segmentation is mainly performed in the detected bounding box (ignoring small differences caused by RoIAlign). This paradigm makes the segmentation performance overly dependent on the bounding box detection performance. However, the core of the instance segmentation task is to distinguish different object instances and segment each instance with a corresponding mask. Bounding boxes can be used for instance identification and roughly locating instances, so bounding boxes can be regarded as tools to assist instance segmentation. If the segmentation is mainly restricted to the inside of the bounding boxes, the performance of instance segmentation will be degraded. Furthermore, since OBBs are sensitive to both position and orientation and are difficult to detect, this dependence makes instance segmentation methods using OBBs more challenging and unstable. 

In this paper, to reduce dependence on bounding box detection performance, we propose OBSeg, an accurate and fast instance segmentation framework that uses box prompt-based segmentation foundation models (BSMs), e.g., Segment Anything Model (SAM) \cite{SAM}. Our motivation is inspired by the powerful segmentation capabilities of BSMs and their excellent box prompt mechanisms. Although there are currently some instance segmentation methods \cite{RSPrompter} using BSMs (e.g., RSPrompter \cite{RSPrompter} improves on SAM), they use HBBs as prompts, introducing many interference regions. Different from BSMs with HBB prompts, OBSeg utilizes more accurate OBBs as prompts, which exhibits better segmentation performance. Specifically, OBSeg first detects OBBs to distinguish instances, identify classes, and provide coarse localization information. Then, it predicts OBB-related segmentation masks. OBSeg uses an OBB detection module \cite{Oriented_R-CNN, Ours, OBB_Detection_TIM_1} for coarse detection, and leverages BSMs to predict OBB-related masks for fine segmentation.

OBSeg improves BSMs in two aspects. First, to handle OBB prompts, OBSeg introduces a novel OBB prompt encoder. Second, since BSMs usually have high computational complexity, we perform knowledge distillation \cite{KD_Survey} on BSMs to improve the practicality. Although there are many methods \cite{EfficientSAM, MobileSAM, EdgeSAM} that use knowledge distillation to introduce lightweight BSMs, the performance gap between the distilled model and the teacher model needs to be further reduced. In OBSeg, a Gaussian smoothing method for teacher model outputs is introduced to further improve distillation performance.

Compared with the ``segmentation within bounding box'' paradigm, OBSeg only uses OBBs as prompts to guide object segmentation, so the segmentation results are less dependent on bounding box detection performance, which effectively improves instance segmentation performance. Moreover, after distillation, OBSeg is more lightweight. The main contributions are summarized as follows.

\begin{enumerate}{}{}

\item We propose OBSeg, an accurate and fast instance segmentation framework using BSMs with OBB prompts. OBSeg outperforms current instance segmentation methods on multiple datasets in terms of instance segmentation accuracy and has competitive inference speed. 

\item A novel OBB prompt encoder is proposed to effectively encode OBB prompts and guide BSMs to generate OBB prompt-related segmentation masks. 

\item A Gaussian smoothing-based knowledge distillation method is introduced to improve the instance segmentation performance of lightweight distilled BSMs.

\end{enumerate}

\section{Related Work}
\subsection{Instance Segmentation with Oriented Bounding Boxes}
Compared with instance segmentation methods using HBBs \cite{Mask_R-CNN, Mask_DINO}, instance segmentation methods using OBBs that have fewer interference regions provide more accurate location information. Most of existing methods build upon the ``segmentation within bounding box'' paradigm and the improved Mask R-CNN \cite{Mask_R-CNN} framework. For example, a Region of Interest (RoI) learner was applied to HBB proposals to generate OBB proposals in ISOP \cite{ISOP}, followed by a head that generated segmentation masks for OBB proposals. Follmann and König \cite{OIS} first generated the final OBBs using a two-stage object detection approach, and then predicted masks within the detected OBBs. Rotated Blend Mask R-CNN \cite{Rotated_Blend} proposed a top-down and bottom-up structure for oriented instance segmentation. Although the above methods effectively improved current instance segmentation performance in remote sensing images, they are overly dependent on OBB detection performance, making predictions more difficult and unstable. Our proposed OBSeg uses OBBs as prompts to guide object segmentation, which effectively reduces this dependence.

\subsection{Box Prompt-Based Segmentation Foundation Models}
Benefiting from pretraining on large-scale data for segmentation tasks, BSMs exhibit powerful segmentation and generalization capabilities. For example, SAM \cite{SAM} was trained on more than 1B segmentation masks from 11M images, showing remarkable segmentation and zero-shot generalization capabilities. A BSM for medical image segmentation is proposed in \cite{Medical_SAM}. To enhance interactive performance, BSMs introduce prompt mechanisms. A representative work is SAM, which took HBBs as prompts and predicted segmentation masks with respect to HBBs. To improve the practicality of BSMs, EfficientSAM \cite{EfficientSAM} and MobileSAM \cite{MobileSAM} used knowledge distillation to introduce lightweight image encoders for edge devices. EdgeSAM \cite{EdgeSAM} performs distillation on the image encoder, prompt encoder, and mask decoder. However, the performance of these distilled BSMs is limited. We introduce a Gaussian smoothing method for teacher model outputs to further improve distillation performance for BSMs.

\begin{figure*}[t]
    \centering
    \centerline{\includegraphics[scale=0.380]{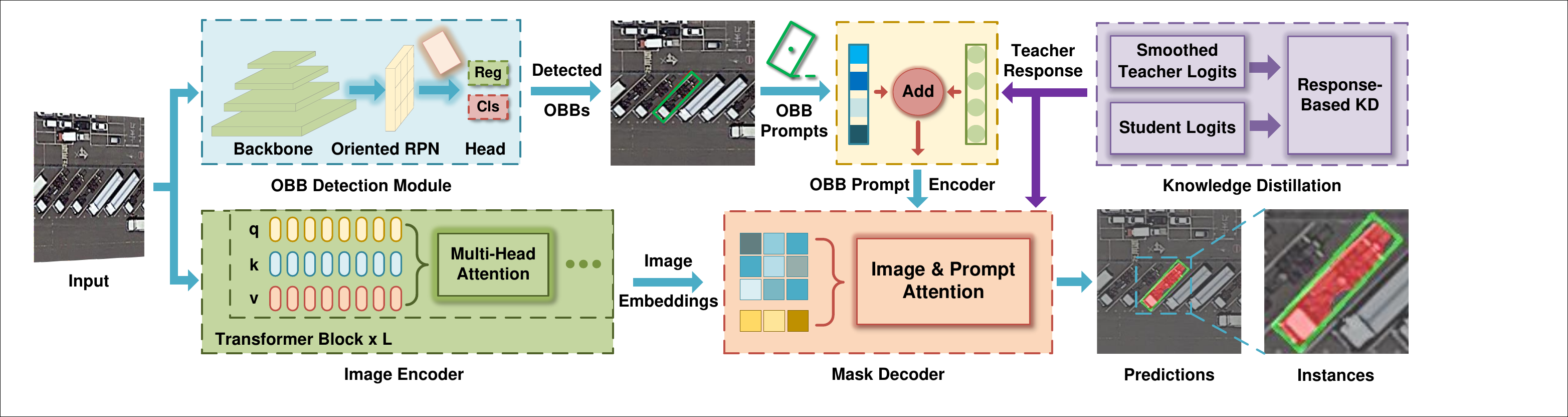}} 
    \caption{Architecture of the proposed OBSeg. It is mainly composed of four parts: an OBB detection module, an image encoder, an OBB prompt encoder, and a mask decoder. OBSeg first detects OBBs to distinguish instances, identify classes, and provide coarse localization information. Then, the mask decoder utilizes the image embeddings generated by the image encoder and the OBB prompt embeddings generated by the OBB prompt encoder to generate segmentation masks. In addition, Gaussian smoothing-based knowledge distillation is performed on the OBB prompt encoder and the mask decoder to make OBSeg more lightweight.}
    \label{fig:Methods_Overview}
\end{figure*}

\subsection{Box Prompt Encoder}
Existing methods usually use HBBs as box prompts for BSMs, such as SAM and EfficientSAM. The current designs of HBB prompt encoders are mainly inspired by SAM. A HBB was represented by two points, i.e., the top-left corner point and the bottom-right corner point. Then, a point was encoded as the sum of a Gaussian positional encoding \cite{Gaussian_PE} of the location and a learned embedding representing the top-left corner or bottom-right corner. Hence, a box was encoded as a combination of the two encoded point embeddings. In remote sensing images, using OBBs helps improve instance segmentation performance. However, to the best of our knowledge, there is no relevant research on OBB prompt encoders yet, which is necessary for BSMs with OBB prompts. 

\subsection{Label Smoothing for Segmentation Networks in Knowledge Distillation}
In knowledge distillation for segmentation tasks, label smoothing helps reduce overfitting and enhances model performance. Inception-v3 \cite{Inception-v3} proposed uniform label smoothing (ULS), i.e., the weighted average of one-hot labels and uniform distributions. Spatially varying label smoothing (SVLS) \cite{SVLS} applied a discrete spatial Gaussian kernel to segmentation labels to smooth one-hot labels. These two label smoothing methods were commonly used in segmentation tasks. For example, Park et al. \cite{PALS} applied ULS and proposed pixel-wise adaptive label smoothing for self-distillation. P-CD \cite{P-CD} utilized SVLS to generate segmentation boundary uncertainty and soft targets. Different from applying SVLS to ground truth (GT) labels, to further improve the generalization capability and instance segmentation performance of the student model, we apply Gaussian smoothing to the segmentation masks of the teacher model, which yields better distillation performance.

\section{Methods}
\subsection{Overview}
The structure of OBSeg is shown in Fig. \ref{fig:Methods_Overview}. It is mainly composed of four parts: an OBB detection module, an image encoder, an OBB prompt encoder, and a mask decoder. Specifically, the OBB detection module is used to detect OBBs that distinguish instances, identify classes, and provide coarse localization information. The image encoder is used to transform input images into high-dimensional feature representations and generate image embeddings. The OBB prompt encoder is responsible for encoding OBB prompts and generating prompt embeddings. Then, the image embeddings and prompt embeddings are combined in the mask decoder to predict OBB prompt-related segmentation masks. Furthermore, Gaussian smoothing-based knowledge distillation is applied to the OBB prompt encoder and mask decoder to make OBSeg more lightweight and practical.

\subsection{Oriented Bounding Box Prompt Encoder}

\begin{figure}[t]
    \centering
    \centerline{\includegraphics[scale=0.380]{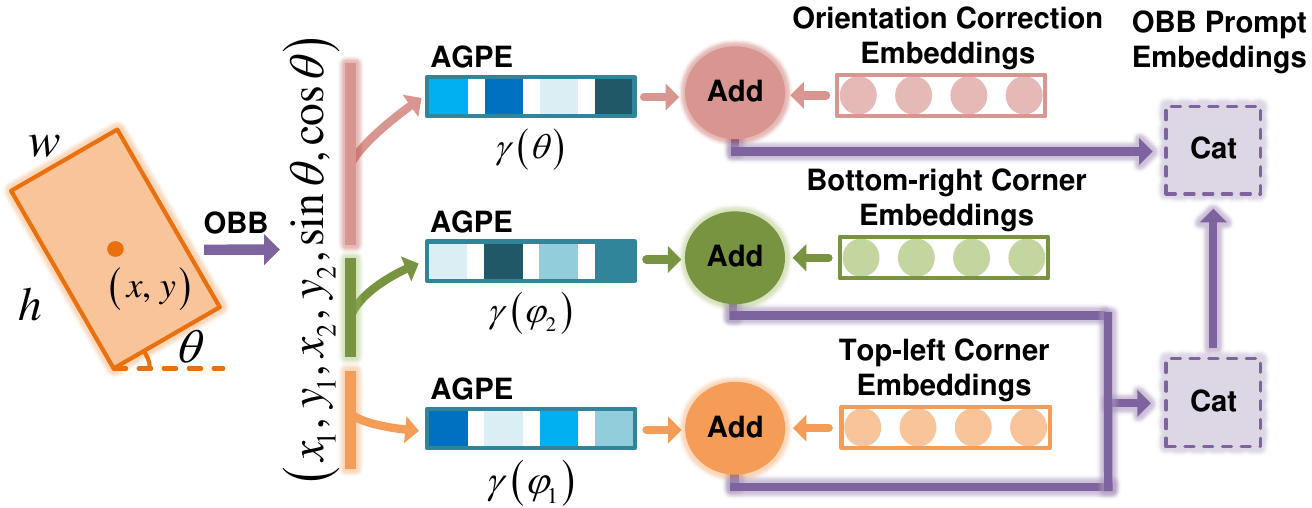}} 
    \caption{Architecture of the proposed OBB prompt encoder. The input is an OBB ($x, y, w, h, \theta$), where $(x, y)$, $w$, $h$ and $\theta$ represent the center point, width, height and orientation, respectively.}
    \label{fig:OBB_Prompt_Encoder}
\end{figure}

The flowchart of the proposed OBB prompt encoder is described in Fig. \ref{fig:OBB_Prompt_Encoder}. The detected OBBs are generated by the OBB detection module. An OBB is first parameterized as ($x_1, y_1, x_2, y_2, \sin\theta, \cos\theta$), where ($x_1, y_1$), ($x_2, y_2$) and ($\sin\theta, \cos\theta$) represent the top-left corner point, bottom-right corner point and orientation, respectively. We further denote
\begin{gather}
    \bm{\varphi }_1 = (x_1,y_1),~~ \bm{\varphi }_2 = (x_2,y_2), ~~ \bm{\theta} = (\sin\theta, \cos\theta), \notag \\
    \bm{\varphi}_1  \in {[0, 1)}^2,~~ \bm{\varphi}_2  \in {[0, 1)}^2,~~ \bm{\theta} \in {[0, 1)}^2, 
\end{gather}
where ($\bm{\varphi}_1, \bm{\varphi}_2$) and $\bm{\theta}$ are normalized pixel coordinates and normalized orientation coordinates, respectively. Considering the difference between position information ($\bm{\varphi}_1, \bm{\varphi}_2$) and orientation information ($\bm{\theta}$), we first encode them separately and then fuse their encoded features.

Coordinate-based multilayer perceptrons (MLPs) take low-dimensional coordinates as input, such as $\bm{\varphi}_1$, $\bm{\varphi}_2$ and $\bm{\theta}$, and they have difficulty learning high-frequency information \cite{Gaussian_PE}. In OBSeg, there are many MLP structures and transformer structures containing MLPs, so such a problem will degrade instance segmentation performance. To alleviate this problem, inspired by Gaussian positional encoding (GPE) for coordinates \cite{Gaussian_PE}, OBSeg introduces adaptive GPE (AGPE) to map the coordinate set ($\bm{\varphi}_1, \bm{\varphi}_2, \bm{\theta}$) into random Fourier features to better learn high-frequency information and enhance the performance of coordinate-based MLPs.

Given a d-dimensional coordinate $\bm{\tau} \in {[0, 1)}^d$, the dense Fourier feature mapping is 
\begin{align}
    \gamma(\bm{\tau}) = [&a_0 \cos(2 \pi \bm{G}_0^T \bm{\tau}), 
                        b_0 \sin(2 \pi \bm{G}_0^T \bm{\tau}), \ldots \notag \\
                        & a_i \cos(2 \pi \bm{G}_i^T \bm{\tau}), 
                        b_i \sin(2 \pi \bm{G}_i^T \bm{\tau}), \ldots]^T,
\end{align}
where $a_i$ and $b_i$ ($i=0, \ldots, \infty$) are Fourier series coefficients, and $\bm{G}_i \in \mathbb{R}^d$ is the corresponding Fourier basis frequency. Since random sparse sampling of Fourier features through an MLP matches the performance of using a dense sampling of Fourier features with the same MLP \cite{Gaussian_PE}, AGPE sets all $a_i$ and $b_i$ to 1. $\bm{G}_i$ is sampled from a Gaussian distribution. Hence, the mapped random Fourier features are 
\begin{align}
    \gamma(\bm{\varphi }_1) &= [\cos(2 \pi G_{\bm{\varphi }_1} \bm{\varphi }_1), \sin(2 \pi G_{\bm{\varphi }_1} \bm{\varphi }_1)]^T, \\
    \gamma(\bm{\varphi }_2) &= [\cos(2 \pi G_{\bm{\varphi }_2} \bm{\varphi }_2), \sin(2 \pi G_{\bm{\varphi }_2} \bm{\varphi }_2)]^T, \\
    \gamma(\bm{\theta}) &= [ \cos(2 \pi G_{\bm{\theta}} \bm{\theta}), \sin(2 \pi G_{\bm{\theta}} \bm{\theta})]^T,
\end{align}
where 
\begin{align}
    G_{\bm{\varphi }_1} &\in \mathbb{R}^{(l_p, 2)} \sim \mathcal{N}(\bm{\rho_}{\bm{\varphi }_1}, \bm{\Sigma}_{\bm{\varphi }_1}),~~\gamma(\bm{\varphi }_1) \in \mathbb{R}^{2l_p}, \\
    G_{\bm{\varphi }_2} &\in \mathbb{R}^{(l_p, 2)} \sim \mathcal{N}(\bm{\rho_}{\bm{\varphi }_2}, \bm{\Sigma}_{\bm{\varphi }_2}), ~~\gamma(\bm{\varphi }_2) \in \mathbb{R}^{2l_p}, \\
    G_{\bm{\theta}} &\in \mathbb{R}^{(l_\theta, 2)} \sim \mathcal{N}(\bm{\rho_}{\bm{\theta}}, \bm{\Sigma}_{\bm{\theta}}) ~~\gamma(\bm{\theta}) \in \mathbb{R}^{2l_\theta}.
\end{align}
The $l_p$ and $l_\theta$ are the lengths of the sampled Fourier basis frequencies. The $(\bm{\rho_}{\bm{\varphi }_1}, \bm{\rho_}{\bm{\varphi }_2}, \bm{\rho_}{\bm{\theta}})$ and ($\bm{\bm{\Sigma}}_{\bm{\varphi }_1}, \bm{\bm{\Sigma}}_{\bm{\varphi }_2}, \bm{\bm{\Sigma}}_{\bm{\theta}}$) represent mean vectors and covariance matrices, respectively. To represent position information ($\gamma(\bm{\varphi }_1)$, $\gamma(\bm{\varphi }_2)$) and orientation information ($\gamma(\bm{\theta})$) in a unified representation space, we set $l_p = l_\theta$. 

In many HBB prompt encoders of BSMs, $(\bm{\rho_}{\bm{\varphi }_1}, \bm{\rho_}{\bm{\varphi }_2}, \bm{\rho_}{\bm{\theta}})$ and ($\bm{\bm{\Sigma}}_{\bm{\varphi }_1}, \bm{\bm{\Sigma}}_{\bm{\varphi }_2}, \bm{\bm{\Sigma}}_{\bm{\theta}}$) are considered as hyperparameters, which leads to difficulty in adapting to changes in datasets and searching optimal parameter settings. In OBSeg, we adaptively learn ($\bm{\bm{\Sigma}}_{\bm{\varphi }_1}, \bm{\bm{\Sigma}}_{\bm{\varphi }_2}, \bm{\bm{\Sigma}}_{\bm{\theta}}$). $(\bm{\rho_}{\bm{\varphi }_1}, \bm{\rho_}{\bm{\varphi }_2}, \bm{\rho_}{\bm{\theta}})$ are fixedly set to 0. Concretely, ($\bm{\bm{\Sigma}}_{\bm{\varphi }_1}, \bm{\bm{\Sigma}}_{\bm{\varphi }_2}, \bm{\bm{\Sigma}}_{\bm{\theta}}$) are the weights of a learnable embedding layer.   

Inspired by SAM, the mapped random Fourier features ($\gamma(\bm{\varphi }_1)$ and $\gamma(\bm{\varphi }_2)$) are summed with two learned embeddings ($\bm{\omega}_1$ and $\bm{\omega}_2$) that represent the top-left corner point and bottom-right corner point in OBSeg, respectively.
\begin{align}
    \mathcal{E}(\bm{\varphi }_1) = \gamma(\bm{\varphi }_1) + \bm{\omega}_1, \\
    \mathcal{E}(\bm{\varphi }_2) = \gamma(\bm{\varphi }_2) + \bm{\omega}_2.
\end{align}
Then, we obtain the encoded feature embeddings $\mathcal{E}(\bm{\varphi}_p)$ with respect to position information, which is given as 
\begin{gather}
    \mathcal{E}(\bm{\varphi}_p) = \mathrm{Concat}[\mathcal{E}(\bm{\varphi }_1),~~ \mathcal{E}(\bm{\varphi }_2)].
\end{gather}

On the other hand, the parameterized orientation representation ($\sin\theta, \cos\theta$) of an OBB suffers from the boundary discontinuity problem \cite{KLD_PAMI}. Due to the periodicity of orientations, at boundary orientations (such as $0^{\circ}$ and $180^{\circ}$), small changes in orientations will result in large jumps in ($\sin\theta, \cos\theta$). To alleviate the boundary discontinuity problem, OBSeg introduces learnable orientation correction embeddings $\bm{\omega}_\theta$. At the orientations where the boundary discontinuity problem occurs, $\bm{\omega}_\theta$ adaptively adjusts the effect of this problem. Hence, the encoded feature embeddings $\mathcal{E}(\bm{\theta})$ with respect to orientation information are 
\begin{gather}
    \mathcal{E}(\bm{\theta}) = \gamma(\bm{\theta}) + \bm{\omega}_\theta.
\end{gather}
Finally, the encoded OBB prompt embeddings $\mathcal{P}_\mathrm{OBB}$ (also called prompt embeddings) are the concatenation of the encoded position feature embeddings and encoded orientation feature embeddings, which is described as
\begin{gather}
    \mathcal{P}_\mathrm{OBB} = \mathrm{Concat}[ \mathcal{E}(\bm{\varphi}_p),~~ \mathcal{E}(\bm{\theta}) ].
\end{gather}

\subsection{Knowledge Distillation}
\begin{figure}[t]
    \centering
    \centerline{\includegraphics[scale=0.445]{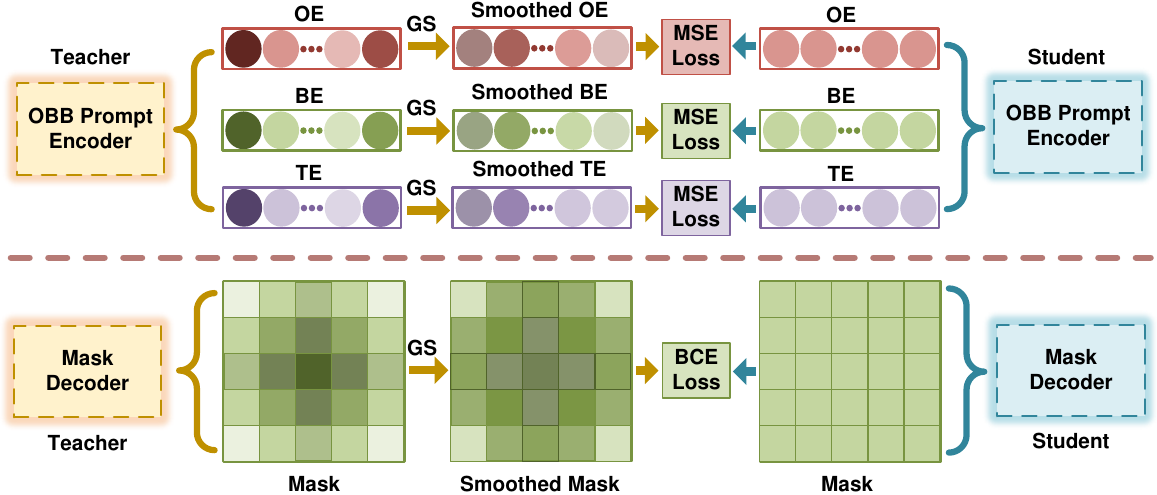}} 
    \caption{The process of knowledge distillation for the OBB prompt encoder and mask decoder. ``TE'', ``BE'' and ``OE'' represent encoded feature embeddings with respect to the top-left point, bottom-right point and orientation of an OBB, respectively. ``GS'' stands for Gaussian smoothing.}
    \label{fig:Knowledge_Distillation}
\end{figure}

BSMs exhibit powerful performance, however they usually have high computational complexity. Although some lightweight BSMs, such as EfficientSAM, MobileSAM, and EdgeSAM, perform knowledge distillation on BSMs, the performance gap between the distilled model and the teacher model is still large. To tackle this, OBSeg introduces a Gaussian smoothing method to enhance distillation performance. Considering the powerful data-fitting capability of the teacher model, the student model only mimics the teacher model and does not learn from GT labels. Furthermore, to further enhance the performance and generalization capability of the student model, Gaussian smoothing is applied to teacher model outputs. While current methods focus on reducing the complexity of the image encoder, we focus on further improving the distillation performance of the OBB prompt encoder and mask decoder. We use a pretrained lightweight image encoder and perform distillation on the OBB prompt encoder and mask decoder, as shown in Fig. \ref{fig:Knowledge_Distillation}.

We first define a one-dimensional Gaussian kernel $\mathbf{G}^1_k (x_i; \sigma)$ with a length of $k$, which is given as
\begin{equation}
    \mathbf{G}^1_k (x_i; \sigma) = \frac{1}{\mathbf{G}^1_s} e^{-[\frac{(x_i-x_c)^2}{2\sigma^2}]},
\end{equation}
where
\begin{equation}
    \mathbf{G}^1_s = \sum_{i}^{} e^{-[\frac{(x_i-x_c)^2}{2\sigma^2}]}, ~~i \in \{0, 1, \ldots, k-1\}.
\end{equation}
The $x_i$ and $x_c$ represent the position of each element and the central position, respectively. $\sigma$ is the standard deviation. Given the encoded feature embeddings of the teacher model with respect to the top-left point ($\mathcal{E}^t(\bm{\varphi }_1) \in \mathbb{R}^{2l}$, $l=l_p=l_\theta$), bottom-right point ($\mathcal{E}^t(\bm{\varphi }_2) \in \mathbb{R}^{2l}$) and orientation ($\mathcal{E}^t(\bm{\theta}) \in \mathbb{R}^{2l}$), we perform Gaussian smoothing by convolving the Gaussian kernel $\mathbf{G}^1_k$ with these encoded embeddings. 
\begin{align}
    \bm{G}^t (\bm{\varphi }_1) &= \mathbf{G}^1_k \circledast \mathcal{E}^t (\bm{\varphi }_1), \\
    \bm{G}^t (\bm{\varphi }_2) &= \mathbf{G}^1_k \circledast \mathcal{E}^t (\bm{\varphi }_2), \\
    \bm{G}^t (\bm{\theta }) &= \mathbf{G}^1_k \circledast \mathcal{E}^t (\bm{\theta }),
\end{align}
where $\circledast$ represents the convolution operation.

Then, the corresponding encoded feature embeddings ($\mathcal{E}^s (\bm{\varphi }_1), \mathcal{E}^s (\bm{\varphi }_2), \mathcal{E}^s (\bm{\theta })$) of the student model mimic the smoothed feature embeddings ($\bm{G}^t (\bm{\varphi }_1), \bm{G}^t (\bm{\varphi }_2), \bm{G}^t (\bm{\theta })$) of the teacher model. The mean squared error (MSE) loss ($\mathcal{L}_\mathrm{prompt}$) is used as the loss function.
\begin{align}
    \mathcal{L}_\mathrm{prompt} &= \frac{1}{2l} {||\bm{G}^t (\bm{\varphi }_1) - \mathcal{E}^s (\bm{\varphi }_1)||}^2_2 \notag \\
                                &+ \frac{1}{2l} {||\bm{G}^t (\bm{\varphi }_2) - \mathcal{E}^s (\bm{\varphi }_2)||}^2_2  \notag \\
                                &+ \frac{1}{2l} {||\bm{G}^t (\bm{\theta }) - \mathcal{E}^s (\bm{\theta })||}^2_2.
\end{align}

\begin{figure*}[t]
    \centering
    \centerline{\includegraphics[scale=0.360]{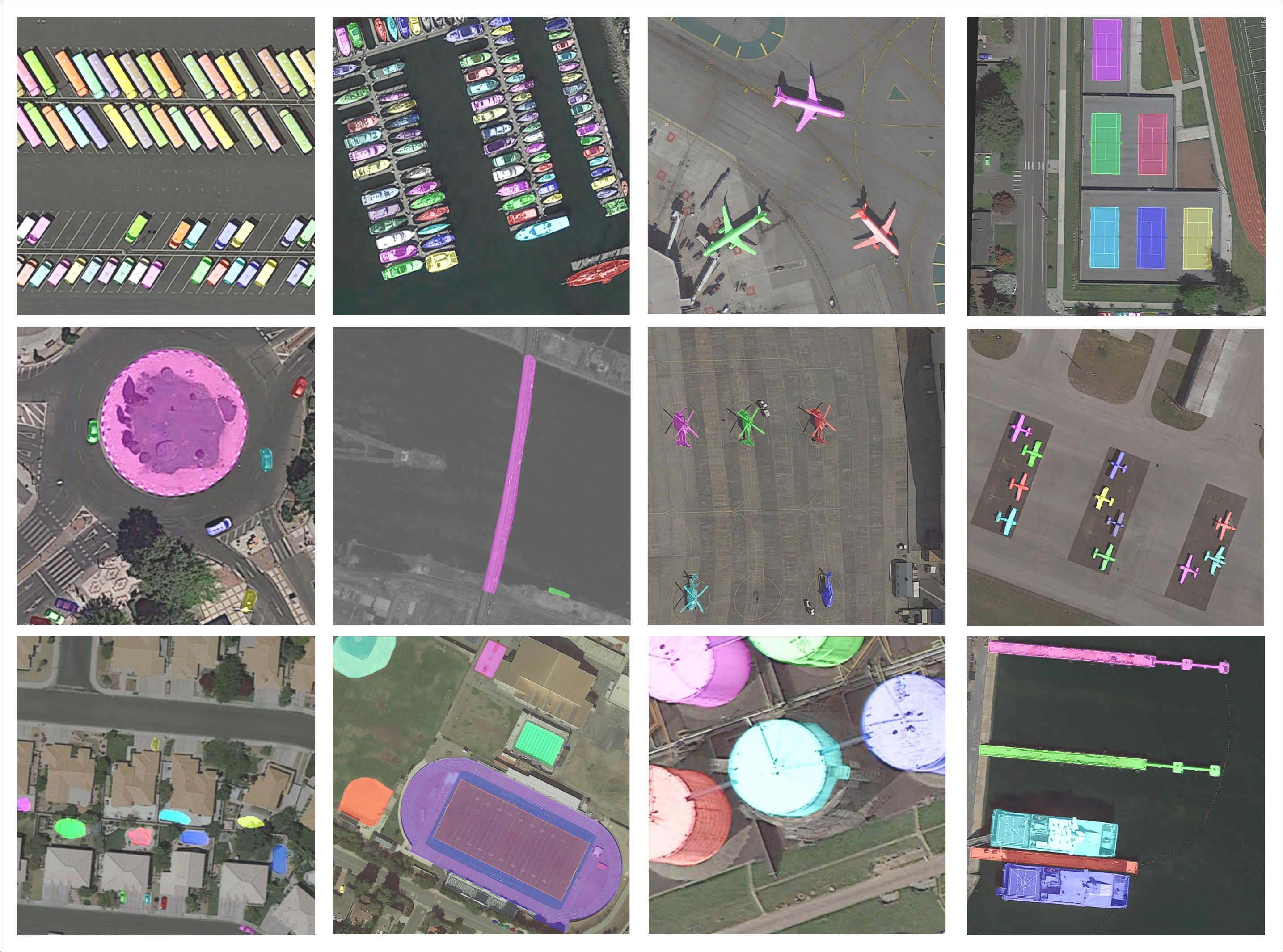}} 
    \caption{Some visualization results of OBSeg's predictions on the iSAID dataset. Segmentation masks of different colors represent different instances.}
    \label{fig:iSAID_2}
\end{figure*}

For knowledge distillation on the mask decoder of a BSM, Gaussian smoothing makes the transition between boundary regions of different classes smoother and improves generalization capability. As the mask is two-dimensional, a two-dimensional Gaussian kernel $\mathbf{G}^2_{k \times k} (u_i, v_j; \delta)$ with a size of $k \times k$ is given as
\begin{equation}
    \mathbf{G}^2_{k \times k} (u_i, v_j; \delta) = \frac{1}{\mathbf{G}^2_s} e^{-[\frac{(u_i-u_c)^2 + (v_j-v_c)^2}{2\delta^2}]},
\end{equation}
where 
\begin{equation}
    \mathbf{G}^2_s = \sum_{i}^{}\sum_{j}^{} e^{-[\frac{(u_i-u_c)^2 + (v_j-v_c)^2}{2\delta^2}]}, ~~i, j \in \{0, 1, \ldots, k-1\}.
\end{equation}
The ($u_i, v_j$) and ($u_c, v_c$) represent the position of each element and the central position, respectively. $\delta$ is the standard deviation. In knowledge distillation for segmentation tasks, unlike current methods \cite{SVLS, P-CD} that apply Gaussian smoothing to GT labels, we apply it to masks generated by the teacher model.

Since the detected OBBs distinguish instances and identify classes, the output mask is only responsible for distinguishing between foreground and background and has a shape of ($H, W$), where $H$ and $W$ represent the height and width of the input image, respectively. Given the output $\bm{M}^t$ of the mask decoder of the teacher model, we first use the sigmoid function $\sigma (\cdot)$ to generate a foreground probability map, and then convolve it with the Gaussian kernel $\mathbf{G}^2_{k \times k}$.
\begin{equation}
    \bm{G} (\bm{M}^t) = \sigma (\bm{M}^t) \circledast \mathbf{G}^2_{k \times k},
\end{equation}
Subsequently, the generated smoothed target $\bm{G} (\bm{M}^t)$ serves as the supervision target for the corresponding probability map ($\sigma (\bm{M}^s)$) of the mask decoder of the student model. We use the binary cross entropy (BCE) loss ($\mathcal{L}_\mathrm{mask}$) as the loss function, which is described as 
\begin{gather}
    \mathcal{L}_\mathrm{mask} = -\frac{1}{HW} \sum_{m^t, m^s}^{} [m^t \log m^s + (1 - m^t) \log(1 - m^s)], \notag \\
    m^t \in \bm{G} (\bm{M}^t), m^s \in \sigma (\bm{M}^s).
\end{gather}
Hence, the final distillation loss ($\mathcal{L}_\mathrm{total}$) is 
\begin{equation}
    \mathcal{L}_\mathrm{total} = \lambda \cdot \mathcal{L}_\mathrm{prompt} + (1 - \lambda) \cdot \mathcal{L}_\mathrm{mask},
\end{equation}
where $\lambda$ is the trade-off factor between the distillation loss of the OBB prompt encoder and that of the mask decoder.

\begin{table*}[h]
    \caption{Comparison results on the iSAID validation set. ``RB Mask R-CNN'' represents Rotated Blend Mask R-CNN (same elsewhere)}
    \footnotesize
    \begin{center}
    \begin{tabular}{m{13em}<{\centering} m{8em}<{\centering} m{4em}<{\centering} m{4em}<{\centering} m{4em}<{\centering} m{4em}<{\centering} m{4em}<{\centering} m{4em}<{\centering} m{4em}<{\centering}}

    \hline
    Method & Backbone & AP & AP\textsubscript{50} & AP\textsubscript{75} & AP\textsubscript{S} & AP\textsubscript{M} & AP\textsubscript{L} & FPS \\ 
    \hline

    SOLO \cite{SOLO}                    & Inception-ResNet & 24.7 & 44.9 & 24.2 & 7.9 & 32.7 & 47.8 & 13.2 \\

    Box2Mask-C \cite{Box2Mask-C}        & ResNet-101-FPN & 26.6 & 50.6 & 23.8 & 10.6 & 33.6 & 47.4 & 11.1 \\

    Luo et al. \cite{Luo_et_al}         & ResNet-50-FPN & 29.4 & 54.5 & 27.8 & 15.5 & 37.8 & 42.0 & 10.7 \\
    
    OIS \cite{OIS}                      & ResNet-50-FPN & 34.1 & 57.2 & 34.2 & 18.5 & 42.1 & 49.9 & 11.0 \\

    PointRend \cite{PointRend}          & ResNet-50-FPN & 35.6 & 59.0 & 37.3 & 20.3 & 44.8 & 52.9 & 15.7 \\
    
    Mask Scoring R-CNN \cite{Mask_Scoring_R-CNN} & ResNet-50-FPN & 35.9 & 57.7 & 38.4 & 20.8 & 44.3 & 51.5 & 9.8 \\

    DCTC \cite{DCTC}                    & DLASeg & 36.3 & 58.2 & 38.0 & 21.2 & 45.2 & 47.3 & \textbf{23.9} \\

    SCNet \cite{SCNet}                  & ResNet-50-FPN & 37.3 & 59.5 & 40.3 & 23.3 & 44.8 & 52.3 & 7.6 \\

    Hybrid Task Cascade \cite{Hybrid_Task_Cascade} & ResNet-50-FPN & 37.4 & 60.2 & 40.1 & 23.5 & 44.6 & 53.5 & 5.8 \\

    Ye et al. \cite{Three_Module_ESWA}  & ResNet-50-FPN & 37.6 & 60.8 & 41.4 & 22.4 & 47.7 & 53.4 & 5.3 \\

    RB Mask R-CNN \cite{Rotated_Blend}  & ResNet-50-FPN & 37.9 & 60.6& 40.9 & 23.0 & 46.8 & 52.7 & 6.1 \\

    R-ARE-Net \cite{ARE-Net}            & DarkNet-19 & 38.0 & 60.8 & 40.7 & 25.2 & 45.9 & 52.7 & - \\

    RB Mask R-CNN \cite{Rotated_Blend}  & ViT-Tiny & 38.5 & 61.5 & 41.4 & 24.0 & 47.3 & 53.3 & 3.8 \\

    ISOP \cite{ISOP}                    & ResNet-101-FPN & 39.4 & 64.2 & 41.7 & 24.9 & 48.6 & 55.4 & 16.6 \\

    Ye et al. \cite{Three_Module_ESWA}  & Swin-Tiny & 40.2 & 65.1 & 44.3 & 25.3 & 49.9 & 57.3 & 2.9 \\

    ISOP \cite{ISOP}                    & ViT-Tiny & 40.3 & 65.2 & 43.9 & 25.6 & 49.4 & 56.8 & 9.1 \\
    
    \hline
    Finetuned MobileSAM \cite{MobileSAM}          & ViT-Tiny & 33.4 & 58.8 & 32.9 & 19.1 & 42.5 & 47.8 & 21.5 \\
    Finetuned EfficientSAM \cite{EfficientSAM}    & ViT-Tiny & 34.8 & 60.0 & 34.0 & 20.3 & 43.7 & 48.6 & 21.5 \\
    Finetuned SAM \cite{SAM}                      & ViT-H & 37.2 & 62.3 & 36.7 & 22.9 & 45.9 & 51.1 & 2.8 \\
    RSPrompter-anchor \cite{RSPrompter}           & ViT-H & 40.9 & 64.0 & 44.2 & 24.3 & 49.7 & 57.0 & 2.7 \\
    RSPrompter-query \cite{RSPrompter}           & ViT-H & 41.6 & 64.4 & 44.6 & 24.8 & 50.2 & 57.6  & 2.7 \\

    \hline
    OBSeg*            & ViT-H & \textbf{45.1} & \textbf{69.4} & \textbf{48.3} & \textbf{28.8} & \textbf{55.1} & \textbf{60.7} & 2.7 \\

    OBSeg$\dagger$    & ViT-Tiny & 43.2 & 64.8 & 45.3 & 25.1 & 51.0 & 59.4 & 20.4 \\

    \rowcolor{gray!20}
    OBSeg             & ViT-Tiny & 43.7 & 65.2 & 46.1 & 25.7 & 51.4 & 59.6 & 20.4 \\

    \hline

\end{tabular}
\end{center}
\label{table:Comparison_iSAID}
\end{table*}

\section{Experiments}
To verify the effectiveness of OBSeg, we conducted experiments on multiple public datasets. Since some datasets do not provide the GT parameters of OBBs, the smallest OBBs of instance masks are used as regression targets.

\subsection{Dataset}
\subsubsection{iSAID} iSAID \cite{iSAID} is a large-scale and densely annotated instance segmentation dataset in aerial images. It contains 2806 high-resolution aerial images and 655,451 object instances across 15 categories. The categories include airplanes, ships, storage tanks, baseball fields, tennis courts, basketball courts, ground track and field, ports, bridges, large vehicles, small vehicles, helicopters, roundabouts, swimming pools, and football fields. The training set, validation set and test set contain 1411, 458 and 937 images, respectively. All images are cropped into $1024 \times 1024$ with a stride of 500. 

\subsubsection{NWPU VHR-10} NWPU VHR-10 instance segmentation dataset \cite{NWPU} is a widely-used dataset in geospatial tasks, which contains 800 high-resolution color and panchromatic infrared images. It has 650 very-high-resolution (VHR) optical images with objects and 150 VHR optical images with pure background. We randomly divide the dataset into the training set and the test set in a ratio of $7\colon3$. 

\subsubsection{PSeg-SSDD} PSeg-SSDD dataset \cite{SSDD} supports ship instance segmentation and ship detection tasks. It contains 1160 SAR images with a resolution ranging from 1 to 15m, and there are various polarizations, resolutions, and scenes. We randomly divide this dataset into a training set of 928 images and a test set of 232 images.

\begin{figure*}[t]
    \centering
    \centerline{\includegraphics[scale=0.247]{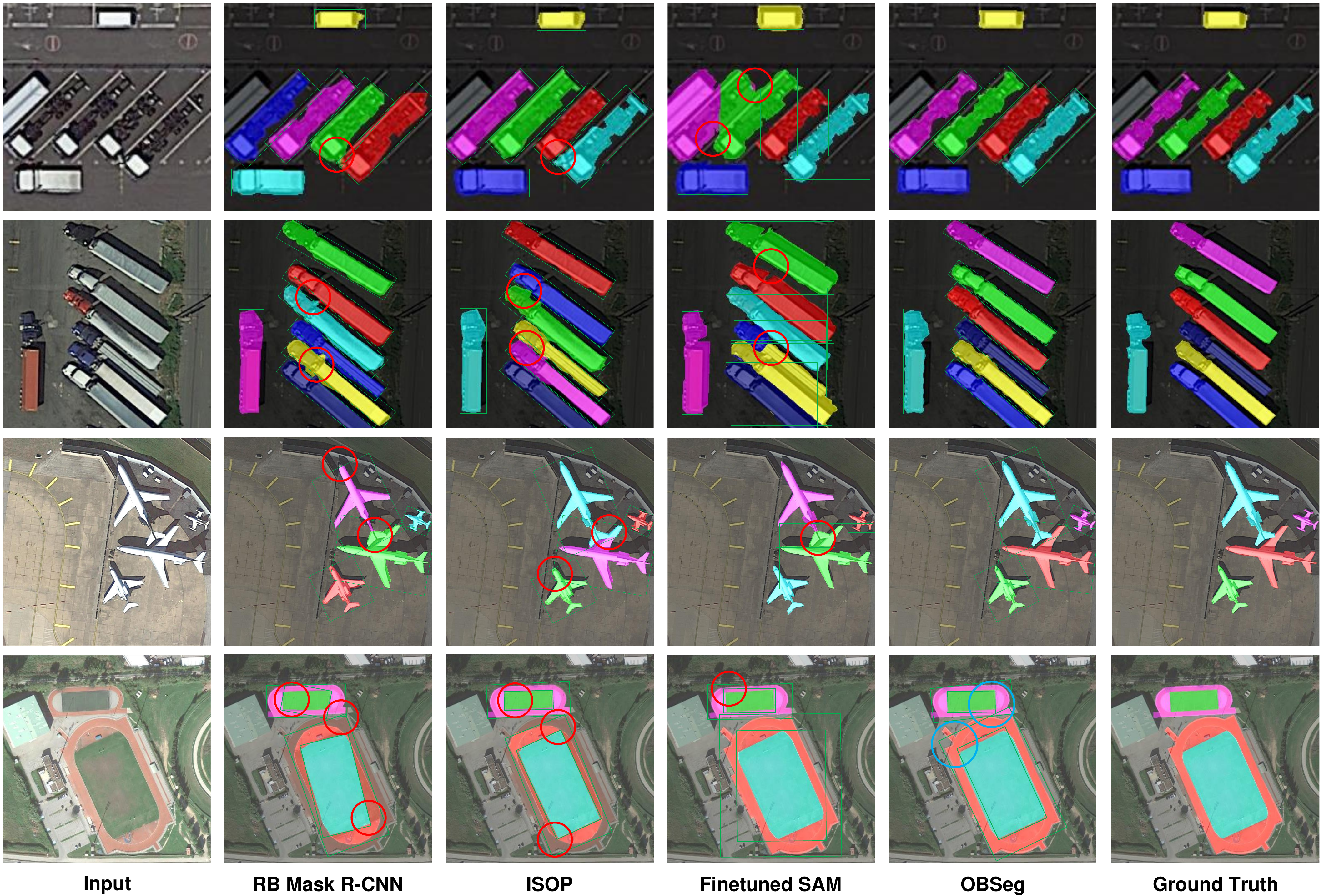}} 
    \caption{Some comparison results on the iSAID dataset. Segmentation masks of different colors represent different instances.}
    \label{fig:iSAID_1}
\end{figure*}

\subsection{Implementation Details}
In this work, OBSeg chooses SAM as the segmentation foundation model. For OBB detection, we directly employ Oriented R-CNN \cite{Oriented_R-CNN}, which is a two-stage OBB detection model with competitive detection performance. The training settings for Oriented R-CNN with ResNet-34 \cite{ResNet} are the same as the original method. In the OBB prompt encoder, the length of the sampled Fourier basis frequencies is 128. For knowledge distillation, the teacher model (called OBSeg*) is based on SAM with the powerful ViT-H (vision transformer \cite{ViT}) image encoder and the proposed OBB prompt encoder. The student model (i.e., OBSeg) directly uses the pretrained lightweight ViT-Tiny image encoder in MobileSAM and the proposed OBB prompt encoder. Their mask decoders and OBB detection modules are the same. During training, the weights of their image encoders are frozen, and the other parts are fine-tuned. Fine-tuning and distillation are performed for 5 and 8 epochs, respectively. The teacher model has the same training loss as SAM and has been trained before distillation. The trade-off factor $\lambda$ for the distillation loss is 0.1. More implementation details can be found in our open-source code. All experiments are conducted on an RTX 3090 GPU. For evaluation, we use the standard evaluation metrics: AP (i.e., AP\textsubscript{50-95}), AP\textsubscript{50}, AP\textsubscript{75}, AP\textsubscript{S}, AP\textsubscript{M} and AP\textsubscript{L}, where ``AP'' means average precision over IoU (Intersection over Union) threshold.

\begin{table*}[t]
    \caption{Comparison results on the NWPU VHR-10 test set}
    \footnotesize
    \begin{center}
    \begin{tabular}{m{13em}<{\centering} m{8em}<{\centering} m{4em}<{\centering} m{4em}<{\centering} m{4em}<{\centering} m{4em}<{\centering} m{4em}<{\centering} m{4em}<{\centering}}

    \hline
    Method & Backbone & AP & AP\textsubscript{50} & AP\textsubscript{75} & AP\textsubscript{S} & AP\textsubscript{M} & AP\textsubscript{L} \\ 
    \hline

    YOLACT \cite{YOLACT}                & ResNet-50-FPN & 43.3 & 77.5 & 44.0 & 23.1 & 40.5 & 54.3 \\

    Mask Scoring R-CNN \cite{Mask_Scoring_R-CNN} & ResNet-50-FPN & 58.4 & 92.1 & 63.2 & 46.7 & 56.9 & 63.5 \\

    PointRend \cite{PointRend}          & ResNet-50-FPN & 61.1 & 90.1 & 64.7 & 52.8 & 59.9 & 61.0 \\

    PointRend \cite{PointRend}          & ViT-Tiny & 61.9 & 91.2 & 65.8 & 53.4 & 61.7 & 61.6 \\

    Hybrid Task Cascade \cite{Hybrid_Task_Cascade} & ResNet-50-FPN & 62.8 & 92.9 & 69.1 & 49.8 & 63.5 & 65.8 \\

    ARE-Net \cite{ARE-Net}              & ResNet-101-FPN & 64.8 & 93.2 & 71.5 & 53.9 & 65.3 & 72.9  \\

    Shi and Zhang \cite{Shi_and_Zhang}  & ResNet-FPN & 65.2 & 94.9 & 72.1 & 49.4 & 65.7 & 71.2 \\

    FB-ISNet \cite{FB-ISNet}            & DLA-BiFPN & 67.0 & 94.5 & 72.0 & - & - & - \\

    DCTC \cite{DCTC}                    & DLASeg & 67.7 & 91.8 & 75.5 & 56.4 & 66.4 & 69.9 \\

    Ye et al. \cite{Three_Module_ESWA}  & Swin-Tiny & 68.2 & - & - & 50.9 & 68.4 & 65.7 \\

    Ye et al. \cite{Three_Module_ESWA}  & ViT-Tiny & 68.2 & - & - & 51.4 & 68.0 & 65.8 \\
    
    \hline
    Finetuned MobileSAM \cite{MobileSAM}          & ViT-Tiny & 60.8 & 91.0 & 63.1 & 51.4 & 61.5 & 60.7 \\
    Finetuned EfficientSAM \cite{EfficientSAM}    & ViT-Tiny & 62.6 & 92.5 & 65.9 & 51.8 & 63.0 & 63.1 \\
    Finetuned SAM \cite{SAM}           & ViT-H & 65.0 & 95.0 & 70.4 & 53.6 & 65.2 & 73.0 \\
    RSPrompter-anchor \cite{RSPrompter}           & ViT-H & 66.1 & 92.7 & 70.6 & - & - & - \\
    RSPrompter-query \cite{RSPrompter}           & ViT-H & 67.5 & 91.7 & 74.8 & - & - & - \\

    \hline
    OBSeg*            & ViT-H & \textbf{71.1} & \textbf{96.7} & \textbf{77.4} & \textbf{59.0} & \textbf{70.5} & \textbf{74.6} \\

    OBSeg$\dagger$    & ViT-Tiny & 69.1 & 94.6 & 74.8 & 57.2 & 68.8 & 72.7 \\

    \rowcolor{gray!20}
    OBSeg             & ViT-Tiny & 69.5 & 95.1 & 75.6 & 57.7 & 69.2 & 73.3 \\

    \hline

\end{tabular}
\end{center}
\label{table:Comparison_NWPU}
\end{table*}

\begin{table*}[t]
    \caption{Comparison results on the PSeg-SSDD test set}
    \footnotesize
    \begin{center}
    \begin{tabular}{m{13em}<{\centering} m{8em}<{\centering} m{4em}<{\centering} m{4em}<{\centering} m{4em}<{\centering} m{4em}<{\centering} m{4em}<{\centering} m{4em}<{\centering}}

    \hline
    Method & Backbone & AP & AP\textsubscript{50} & AP\textsubscript{75} & AP\textsubscript{S} & AP\textsubscript{M} & AP\textsubscript{L} \\ 
    \hline

    YOLACT \cite{YOLACT}                & ResNet-50-FPN & 57.8 & 91.4 & 70.9 & 58.4 & 56.5 & 58.6 \\

    SA R-CNN \cite{SA_R-CNN}            & ResNet50-FPN & 59.4 & 90.4 & 77.6 & 63.3 & 62.5 & 47.7 \\

    EMIN \cite{EMIN}                    & - & 61.7 & 94.3 & 76.8 & 62.1 & 61.3 & 40.1 \\

    FL-CSE-ROIE \cite{FL-CSE-ROIE}      & ResNet-101-FPN & 62.6 & 93.7 & 78.3 & 63.3 & 61.2 & 75.0 \\

    MAI-SE-Net \cite{MAI-SE-Net}        & ResNet-101-FPN & 63.0 & 94.4 & 77.6 & 63.3 & 62.5 & 47.7 \\

    LFG-Net \cite{LFG-Net}              & ResNeXt-64 & 64.2 & 95.0 & 81.1 & 63.1 & 68.2 & 43.1 \\

    Mask Scoring R-CNN \cite{Mask_Scoring_R-CNN} & ResNet-50-FPN & 65.4 & 94.1 & 81.9 & 66.4 & 62.0 & 40.5 \\

    PointRend \cite{PointRend}          & ResNet-50-FPN & 65.6 & 94.5 & 82.3 & 67.0 & 62.2 & 16.8 \\

    Hybrid Task Cascade \cite{Hybrid_Task_Cascade} & ResNet-50-FPN & 66.1 & 95.4 & 83.2 & 66.2 & 68.9 & 29.4 \\

    PointRend \cite{PointRend}          & ViT-Tiny & 66.2 & 95.2 & 83.3 & 67.7 & 64.0 & 20.4 \\

    DCTC \cite{DCTC}                    & DLASeg & 67.4 & 93.9 & 83.8 & 64.4 & \textbf{76.6} & \textbf{80.2} \\

    Shi and Zhang \cite{Shi_and_Zhang}  & ResNet-FPN & 67.8 & \textbf{97.5} & 85.0 & 67.7 & 69.1 & 58.9 \\

    Ye et al. \cite{Three_Module_ESWA}  & Swin-Tiny & 68.4 & - & - & 69.8 & 63.7 & 41.0 \\

    Ye et al. \cite{Three_Module_ESWA}  & ViT-Tiny & 68.5 & - & - & 69.4 & 64.6 & 42.5 \\
    
    \hline
    Finetuned MobileSAM \cite{MobileSAM}          & ViT-Tiny & 62.9 & 93.6 & 77.1 & 64.0 & 64.8 & 49.9 \\
    Finetuned EfficientSAM \cite{EfficientSAM}    & ViT-Tiny & 63.5 & 94.5 & 79.6 & 64.3 & 65.2 & 54.3 \\
    Finetuned SAM \cite{SAM}                      & ViT-H & 66.5 & 95.0 & 83.3 & 64.4 & 66.7 & 58.2 \\
    RSPrompter-anchor \cite{RSPrompter}           & ViT-H & 66.8 & 94.7 & 84.0 & - & - & - \\
    RSPrompter-query \cite{RSPrompter}           & ViT-H & 67.3 & 95.6 & 84.3 & - & - & - \\

    \hline
    OBSeg*            & ViT-H & \textbf{71.8} & 97.0 & \textbf{87.5} & \textbf{71.2} & 75.3 & 72.4 \\

    OBSeg$\dagger$    & ViT-Tiny & 69.3 & 95.7 & 85.0 & 68.8 & 72.5 & 66.1 \\

    \rowcolor{gray!20}
    OBSeg             & ViT-Tiny & 69.8 & 96.1 & 85.4 & 69.4 & 73.1 & 66.6 \\

    \hline

\end{tabular}
\end{center}
\label{table:Comparison_SSDD}
\vspace{-0.2cm}
\end{table*}

\subsection{Comparisons with the State-of-the-Art}
\subsubsection{Comparison Experiments on the iSAID Dataset}
To evaluate the effectiveness of OBSeg, we compare OBSeg with current instance segmentation methods using OBBs on the iSAID dataset. We also include some instance segmentation methods using HBBs (e.g., SAM after finetuned on this dataset) and other SAM-based methods using HBBs (e.g., RSPrompter) for comparison. For fairness, SAM and the teacher model OBSeg* use ViT-H as the image encoder, and OBSeg uses ViT-Tiny as the image encoder. We also replace the backbones of some non-BSM-based methods with pretrained ViT-Tiny. SAM, MobileSAM, and EfficientSAM apply powerful YOLOv7 \cite{YOLOv7} as the HBB detector. ``$\dagger$'' indicates that Gaussian smoothing is not applied during distillation, similar to its use in current lightweight BSMs. The inference is conducted on the RTX 3090 GPU, and the inference speed of some models is based on the results we reproduced.  

As presented in Table \ref{table:Comparison_iSAID}, OBSeg* and OBSeg significantly outperform other methods, achieving 45.1\% AP and 43.7\% AP, respectively. Compared with instance segmentation methods using HBBs, methods using OBBs perform better since OBBs are more suitable for segmenting densely packed objects. Significant accuracy improvements over fine-tuned SAM series and other SAM-based methods using HBBs also indicate the benefit of using OBBs. In addition, OBSeg is less dependent on OBB detection performance and outperforms existing ``segmentation within bounding box'' methods by a large margin. Fig. \ref{fig:iSAID_1} shows some instance segmentation visualization results of OBSeg and current methods. For dense objects with oblique orientations, the segmentation results of methods using HBBs incorrectly include other instances, reducing the instance segmentation performance. Furthermore, when predicted OBBs significantly deviate from GT OBBs, the segmentation performance of existing ``segmentation within bounding box'' methods is substantially affected. Based on the OBB prompt mechanism, OBSeg still accurately segments instances when the OBB predictions are inaccurate.

Compared with other BSMs, OBSeg is very lightweight. Even though SAM has a powerful ViT-H image encoder, its performance is still inferior to OBSeg. Among BSMs of the same lightweight level, OBSeg achieves higher accuracy and a competitive inference speed of 20.4 FPS, which is also better than most methods not based on BSMs. This demonstrates that OBSeg is more efficient than current instance segmentation methods. Moreover, compared with OBSeg*, the notable inference speed improvement and minor accuracy loss of OBSeg also demonstrate the effectiveness of the proposed knowledge distillation method. Gaussian smoothing further enhances the distillation performance and generalization capability. More instance segmentation results of OBSeg on the iSAID dataset are displayed in Fig. \ref{fig:iSAID_2}.

\subsubsection{Comparison Experiments on the NWPU VHR-10 Dataset}
We further verify the instance segmentation performance of OBSeg on the NWPU VHR-10 dataset. For fairness, we compare instance segmentation methods using HBBs, methods using OBBs, and BSM-based methods. As in the setting on the iSAID dataset, the HBB detectors in the BSM series all use the powerful YOLOv7. We also replace the backbones of some non-BSM-based methods with ViT-Tiny.

The comparison results in Table \ref{table:Comparison_NWPU} show that OBSeg* and OBSeg achieve the best instance segmentation performance in all accuracy evaluation metrics. The superiority of OBSeg demonstrates the effectiveness of using OBBs as detection bounding boxes and the paradigm of OBB prompt-based instance segmentation. Compared with the teacher model OBSeg*, the proposed distillation method significantly improves the instance segmentation inference speed with less accuracy loss. In addition, compared with OBSeg$\dagger$, Gaussian smoothing further improves the distillation performance. 

\subsubsection{Comparison Experiments on the PSeg-SSDD Dataset}
Similar to the experimental settings of the iSAID and NWPU VHR-10 datasets, we compare methods that use HBBs, OBBs, and those based on BSMs for instance segmentation. 

As presented in Table \ref{table:Comparison_SSDD}, similar to the comparison experiment results on the iSAID and NWPU VHR-10 datasets, OBSeg* and OBSeg show their superior performance. Before distillation, the teacher model OBSeg* achieves an accuracy of 71.8\% AP. After distillation, OBSeg greatly retains the model's performance, achieving an instance segmentation accuracy of 69.8\% AP, which is superior to current methods. 

\subsection{Ablation Studies}
To evaluate the effectiveness of the components of OBSeg, a series of ablation experiments are conducted on the iSAID dataset. We use the most commonly used AP\textsubscript{50} as the metric.

\subsubsection{Oriented Bounding Box Prompt Encoder}
The OBB prompt encoder introduces positional encoding and specific embeddings to encode OBBs. In Table \ref{table:OBB_prompt_encoder_experiments}, ``OCE'', ``TCE'' and ``BCE'' represent the specific orientation correction embeddings, top-left corner embeddings and bottom-right corner embeddings, respectively. Compared with GPE, AGPE encodes location information better by adaptively learning parameters of Gaussian distributions. Moreover, specific embeddings further improve instance segmentation performance by 0.97\% AP\textsubscript{50}. Orientation correction embeddings alleviate the periodicity problem of orientation.

\subsubsection{Knowledge Distillation}
We compare the performance of OBSeg learning from the teacher model and GT labels, and test the impact of different label smoothing methods. ``T'' represents learning from the teacher model. The smoothing factor in ULS is 0.1. For distillation on the mask decoder, Gaussian smoothing (GS) has the same settings as SVLS, with a Gaussian kernel size of $3 \times 3$ and a standard deviation of 1.0. For distillation on the OBB prompt encoder, GS has a Gaussian kernel length of 3 and a standard deviation of 0.3. The weight factor of the teacher model and GT labels is 0.1. ``GS$^{-}$'' represents distillation only for the mask decoder.

The experimental results in Table \ref{table:KD_comparison} demonstrate that learning from the teacher model has higher performance than learning from GT labels (learning only from GT labels means no knowledge distillation is performed), which is mainly thanks to the powerful data-fitting capabilities of BSMs. Furthermore, adding distillation for the OBB prompt encoder effectively enhances performance. Moreover, compared with distillation methods that are not based on Gaussian smoothing in existing lightweight distilled BSMs (e.g., EfficientSAM and MobileSAM), Gaussian smoothing in OBSeg further improves distillation performance and is superior to ULS. In addition, we compare the impact of different parameter settings in Gaussian smoothing on instance segmentation performance in Table \ref{table:GS_parameters}. When $k=5$, $\sigma=0.3$, and $\delta=1.0$, the highest instance segmentation accuracy of 65.24\% AP\textsubscript{50} is achieved.

\begin{table}[t]
    \caption{Performance of each component of the OBB prompt encoder}
    \footnotesize
    \begin{center}
    \begin{tabular}{m{4em}<{\centering}m{4em}<{\centering} m{5em}<{\centering} m{4em}<{\centering} m{4em}<{\centering}} 

    \hline
    GPE & AGPE & TCE+BCE & OCE & AP\textsubscript{50} \\
    \hline
    \checkmark &  &  &  & 64.00 \\
      & \checkmark &  &  & 64.27 \\
      & \checkmark & \checkmark &  & 64.86 \\
    \rowcolor{gray!20}
      & \checkmark & \checkmark & \checkmark & \textbf{65.24} \\
    \hline

\end{tabular}
\end{center}
\label{table:OBB_prompt_encoder_experiments}
\end{table}

\begin{table}[t]
    \caption{Comparison of different learning targets and label smoothing methods in knowledge distillation}
    \footnotesize
    \begin{center}
    \begin{tabular}{m{3.2em}<{\centering} m{3em}<{\centering}m{3em}<{\centering} m{1.8em}<{\centering} m{3em}<{\centering} m{3em}<{\centering} m{2.8em}<{\centering}} 

    \hline
    GT+ULS & GT+GS & T+GT & T & T+ULS & T+GS$^{-}$ & T+GS \\
    64.23 & 64.44 & 64.73 & 64.79 & 64.87 & 64.91 & \textbf{65.16} \\
    \hline

\end{tabular}
\end{center}
\label{table:KD_comparison}
\end{table}

\begin{table}[t]
    \caption{Impact of parameter settings in Gaussian smoothing}
    \footnotesize
    \begin{center}
    \begin{tabular}{m{4em}<{\centering} m{4em}<{\centering} m{4em}<{\centering} m{4em}<{\centering} m{4em}<{\centering}}

    \hline
    \multirow{2}{*}{} & \multicolumn{2}{c}{$\sigma=0.3$} & \multicolumn{2}{c}{$\sigma=0.5$} \\
    \cline{2-5}
     & $\delta=0.5$ & $\delta=1.0$ & $\delta=0.5$ & $\delta=1.0$ \\
    \hline
    $k=3$ & 65.16 & 65.16 & 65.12 & 65.09 \\
    $k=5$ & 65.14 & \textbf{65.24} & 65.10 & 65.13 \\
    $k=7$ & 65.11 & 65.17 & 65.07 & 65.06 \\
    \hline

    \end{tabular}
    \end{center}
    \label{table:GS_parameters}
  \end{table}

  \begin{table}[t]
    \caption{Performance comparison of OBB and HBB detectors}
    \footnotesize
    \begin{center}
    \begin{tabular}{m{6em}<{\centering} m{6em}<{\centering} m{6.2em}<{\centering} m{5em}<{\centering}} 

    \hline
    \multirow{2}{*}{YOLOv7 \cite{YOLOv7}} & Oriented & Oriented & \multirow{2}{*}{S2ANet \cite{S2ANet}} \\
    & R-CNN \cite{Oriented_R-CNN} & Reppoints \cite{Oriented_Reppoints} &  \\
    58.81 & 65.24 & \textbf{65.37} & 65.09 \\
    \hline

\end{tabular}
\end{center}
\label{table:OBB_comparison}
\end{table}

\subsubsection{Oriented Bounding Box Detection}
To explore the impact of the OBB detection module on the instance segmentation performance, we compare Oriented R-CNN used in OBSeg with some other OBB detection methods, such as Oriented Reppoints \cite{Oriented_Reppoints} and S2ANet \cite{S2ANet}. In addition, the HBB detection method YOLOv7 is also added for fair comparison. As shown in Table \ref{table:OBB_comparison}, the effective use of OBBs significantly improves instance segmentation accuracy, and OBSeg is not sensitive to specific OBB detection methods.

\section{Discussion}
\subsubsection{Comparison with Other Methods Using BSMs}
Currently, there are some works that use BSMs, such as SAM, for instance segmentation tasks. These methods mainly aim to leverage the powerful segmentation capabilities of BSMs, which are acquired after training on large-scale datasets. However, these methods rely on HBB prompt input, which introduces many interference regions when performing instance segmentation on objects densely packed in multiple orientations, thereby reducing the segmentation performance. Different from these methods, our proposed OBSeg uses more accurate OBBs as prompts and achieves better instance segmentation performance.

Moreover, we introduce Gaussian smoothing-based knowledge distillation to make OBSeg more lightweight. Experimental results on iSAID, NWPU VHR-10, and PSeg-SSDD datasets demonstrate that OBSeg is superior to current BSM-based methods in terms of instance segmentation accuracy and has competitive inference speed. 

\subsubsection{Analysis of Inference Accuracy and Speed}
Compared with methods using HBBs or those not based on BSMs, OBSeg uses BSMs with OBB prompts, which have fewer interference regions and are less dependent on bounding box detection performance. This enables OBSeg to achieve higher instance segmentation accuracy on multiple datasets.

In addition, since BSMs have powerful segmentation capabilities but slow inference speed, we have applied model lightweighting. To maintain high segmentation performance, we propose a knowledge distillation method based on Gaussian smoothing. Compared with existing methods, the lightweight distilled OBSeg has competitive inference speed while maintaining high segmentation accuracy.

\subsubsection{Limitations and Future Work}
Although OBSeg achieves better instance segmentation performance than existing methods, it also has some limitations. One limitation of this work is that the instance segmentation performance of OBSeg is not robust to objects with very large aspect ratios or extreme scales. This is mainly because these types of objects are in the minority in the dataset, and the model has insufficient feature recognition capabilities for objects with these characteristics. To address this type of long-tail distribution problem, future work will build upon the BSM framework by adopting class-balanced sampling strategies, extreme-scale feature learning, and adaptive adjustment of loss functions.

\section{Conclusions}
This paper proposes OBSeg, an accurate and fast instance segmentation framework using BSMs with OBB prompts in remote sensing images. OBSeg introduces a novel OBB prompt encoder, and we perform knowledge distillation to make OBSeg more lightweight and practical. Specifically, a Gaussian smoothing-based distillation method is introduced to improve the instance segmentation performance of lightweight distilled BSMs. Thanks to OBB prompts, OBSeg is less dependent on bounding box detection performance and has fewer interference regions than instance segmentation methods using HBBs or HBB prompts. Experiments on multiple public datasets, such as iSAID and NWPU VHR-10 datasets, demonstrate that OBSeg outperforms current instance segmentation methods in terms of instance segmentation accuracy and has competitive inference speed. One limitation of this work is that the instance segmentation performance of OBSeg is not robust to objects with very large aspect ratios or extreme scales. Future work will focus on improvements in data, extreme-scale feature learning, and loss functions.

\bibliographystyle{IEEEtran}

\bibliography{IEEEabrv, root}

\begin{thebibliography}{10}
\providecommand{\url}[1]{#1}
\csname url@rmstyle\endcsname
\providecommand{\newblock}{\relax}
\providecommand{\bibinfo}[2]{#2}
\providecommand\BIBentrySTDinterwordspacing{\spaceskip=0pt\relax}
\providecommand\BIBentryALTinterwordstretchfactor{4}
\providecommand\BIBentryALTinterwordspacing{\spaceskip=\fontdimen2\font plus
\BIBentryALTinterwordstretchfactor\fontdimen3\font minus \fontdimen4\font\relax}
\providecommand\BIBforeignlanguage[2]{{%
\expandafter\ifx\csname l@#1\endcsname\relax
\typeout{** WARNING: IEEEtran.bst: No hyphenation pattern has been}%
\typeout{** loaded for the language `#1'. Using the pattern for}%
\typeout{** the default language instead.}%
\else
\language=\csname l@#1\endcsname
\fi
#2}}

\bibitem{Jstars_Orientated_Silhouette_Matching}
Z.~Huang and R.~Li, ``Orientated silhouette matching for single-shot ship instance segmentation,'' \emph{IEEE J. Sel. Top. Appl. Earth Observ. Remote Sens.}, vol.~15, pp. 463--477, 2022.

\bibitem{Jstars_SAR_Ship_Instance}
F.~Gao, X.~Han, J.~Wang, J.~Sun, A.~Hussain, and H.~Zhou, ``Sar ship instance segmentation with dynamic key points information enhancement,'' \emph{IEEE J. Sel. Top. Appl. Earth Observ. Remote Sens.}, vol.~17, pp. 11\,365--11\,385, 2024.

\bibitem{Jstars_An_Edge_Aware}
Y.~Liu, T.~Zhang, Y.~Huang, and F.~Shi, ``An edge-aware multitask network based on cnn and transformer backbone for farmland instance segmentation,'' \emph{IEEE J. Sel. Top. Appl. Earth Observ. Remote Sens.}, vol.~17, pp. 13\,765--13\,779, 2024.

\bibitem{HBB_Det_TIM}
C.~Zhang, T.~Liu, J.~Xiao, K.-M. Lam, and Q.~Wang, ``Boosting object detectors via strong-classification weak-localization pretraining in remote sensing imagery,'' \emph{IEEE Trans. Instrum. Meas.}, vol.~72, pp. 1--20, 2023.

\bibitem{HBB_Det_TIM_2}
W.~Zhao, Y.~Kang, H.~Chen, Z.~Zhao, Z.~Zhao, and Y.~Zhai, ``Adaptively attentional feature fusion oriented to multiscale object detection in remote sensing images,'' \emph{IEEE Trans. Instrum. Meas.}, vol.~72, pp. 1--11, 2023.

\bibitem{Rotated_Blend}
Z.~Zhang and J.~Du, ``Accurate oriented instance segmentation in aerial images,'' in \emph{Int. Conf. Image Graphics}, 2021, pp. 160--170.

\bibitem{OIS}
P.~Follmann and R.~König, ``Oriented boxes for accurate instance segmentation,'' in \emph{arXiv preprint arXiv:1911.07732}, 2020.

\bibitem{Towards_Robust_Part}
Y.~Feng, B.~Yang, X.~Li, C.-W. Fu, R.~Cao, K.~Chen, Q.~Dou, M.~Wei, Y.-H. Liu, and P.-A. Heng, ``Towards robust part-aware instance segmentation for industrial bin picking,'' in \emph{IEEE Int. Conf. Robot. Autom.}, 2022, pp. 405--411.

\bibitem{ISOP}
T.~Pan, J.~Ding, J.~Wang, W.~Yang, and G.-S. Xia, ``Instance segmentation with oriented proposals for aerial images,'' in \emph{IEEE Int. Geosci. Remote Sens. symp.}, 2020, pp. 988--991.

\bibitem{SAM}
A.~Kirillov, E.~Mintun, N.~Ravi, H.~Mao, C.~Rolland, L.~Gustafson, T.~Xiao, S.~Whitehead, A.~C. Berg, W.-Y. Lo, P.~Dollar, and R.~Girshick, ``Segment anything,'' in \emph{IEEE Int. Conf. Comput. Vis.}, October 2023, pp. 4015--4026.

\bibitem{RSPrompter}
K.~Chen, C.~Liu, H.~Chen, H.~Zhang, W.~Li, Z.~Zou, and Z.~Shi, ``Rsprompter: Learning to prompt for remote sensing instance segmentation based on visual foundation model,'' \emph{IEEE Trans. Geosci. Remote Sensing}, vol.~62, pp. 1--17, 2024.

\bibitem{Oriented_R-CNN}
X.~Xie, G.~Cheng, J.~Wang, X.~Yao, and J.~Han, ``Oriented r-cnn for object detection,'' in \emph{IEEE Int. Conf. Comput. Vis.}, October 2021, pp. 3520--3529.

\bibitem{Ours}
Z.~Zhou, Y.~Ma, J.~Fan, Z.~Liu, F.~Jing, and M.~Tan, ``Linear gaussian bounding box representation and ring-shaped rotated convolution for oriented object detection,'' \emph{Pattern Recognit.}, vol. 155, p. 110677, 2024.

\bibitem{OBB_Detection_TIM_1}
P.~Sun, Y.~Zheng, W.~Wu, W.~Xu, S.~Bai, and X.~Lu, ``Learning critical features for arbitrary-oriented object detection in remote-sensing optical images,'' \emph{IEEE Trans. Instrum. Meas.}, vol.~73, pp. 1--12, 2024.

\bibitem{KD_Survey}
J.~Gou, B.~Yu, S.~J. Maybank, and D.~Tao, ``Knowledge distillation: A survey,'' \emph{Int. J. Comput. Vis.}, vol. 129, pp. 1789--1819, 2021.

\bibitem{EfficientSAM}
Y.~Xiong, B.~Varadarajan, L.~Wu, X.~Xiang, F.~Xiao, C.~Zhu, X.~Dai, D.~Wang, F.~Sun, F.~Iandola, R.~Krishnamoorthi, and V.~Chandra, ``Efficientsam: Leveraged masked image pretraining for efficient segment anything,'' in \emph{IEEE Int. Conf. Comput. Vis. Pattern Recognit.}, June 2024, pp. 16\,111--16\,121.

\bibitem{MobileSAM}
C.~Zhang, D.~Han, Y.~Qiao, J.~U. Kim, S.-H. Bae, S.~Lee, and C.~S. Hong, ``Faster segment anything: Towards lightweight sam for mobile applications,'' in \emph{arXiv preprint arXiv:2306.14289}, 2023.

\bibitem{EdgeSAM}
C.~Zhou, X.~Li, C.~C. Loy, and B.~Dai, ``Edgesam: Prompt-in-the-loop distillation for on-device deployment of sam,'' in \emph{arXiv preprint arXiv:2312.06660}, 2023.

\bibitem{Mask_R-CNN}
K.~He, G.~Gkioxari, P.~Dollar, and R.~Girshick, ``Mask r-cnn,'' in \emph{IEEE Int. Conf. Comput. Vis.}, Oct 2017.

\bibitem{Mask_DINO}
F.~Li, H.~Zhang, H.~Xu, S.~Liu, L.~Zhang, L.~M. Ni, and H.-Y. Shum, ``Mask dino: Towards a unified transformer-based framework for object detection and segmentation,'' in \emph{IEEE Int. Conf. Comput. Vis. Pattern Recognit.}, June 2023, pp. 3041--3050.

\bibitem{Medical_SAM}
Y.~Huang, X.~Yang, L.~Liu, H.~Zhou, A.~Chang, X.~Zhou, R.~Chen, J.~Yu, J.~Chen, C.~Chen, S.~Liu, H.~Chi, X.~Hu, K.~Yue, L.~Li, V.~Grau, D.-P. Fan, F.~Dong, and D.~Ni, ``Segment anything model for medical images?'' \emph{Med. Image Anal.}, vol.~92, p. 103061, 2024.

\bibitem{Gaussian_PE}
M.~Tancik, P.~Srinivasan, B.~Mildenhall, S.~Fridovich-Keil, N.~Raghavan, U.~Singhal, R.~Ramamoorthi, J.~Barron, and R.~Ng, ``Fourier features let networks learn high frequency functions in low dimensional domains,'' in \emph{Adv. Neural Inf. Process. Syst.}, vol.~33, 2020, pp. 7537--7547.

\bibitem{Inception-v3}
C.~Szegedy, V.~Vanhoucke, S.~Ioffe, J.~Shlens, and Z.~Wojna, ``Rethinking the inception architecture for computer vision,'' in \emph{IEEE Int. Conf. Comput. Vis. Pattern Recognit.}, June 2016.

\bibitem{SVLS}
M.~Islam and B.~Glocker, ``Spatially varying label smoothing: Capturing uncertainty from expert annotations,'' in \emph{Inf. Process. Med. Imag.}, 2021, pp. 677--688.

\bibitem{PALS}
S.~Park, J.~Kim, and Y.~S. Heo, ``Semantic segmentation using pixel-wise adaptive label smoothing via self-knowledge distillation for limited labeling data,'' \emph{Sensors}, vol.~22, no.~7, 2022.

\bibitem{P-CD}
M.~Islam, L.~Seenivasan, S.~P. Sharan, V.~K. Viekash, B.~Gupta, B.~Glocker, and H.~Ren, ``Paced-curriculum distillation with prediction and label uncertainty for image segmentation,'' in \emph{Int. J. Comput. Assisted Radiology surg.}, vol.~18, 2023, pp. 1875--1883.

\bibitem{KLD_PAMI}
X.~Yang, G.~Zhang, X.~Yang, Y.~Zhou, W.~Wang, J.~Tang, T.~He, and J.~Yan, ``Detecting rotated objects as gaussian distributions and its 3-d generalization,'' \emph{IEEE Trans. Pattern Anal. Mach. Intell.}, vol.~45, no.~4, pp. 4335--4354, 2023.

\bibitem{SOLO}
X.~Wang, T.~Kong, C.~Shen, Y.~Jiang, and L.~Li, ``Solo: Segmenting objects by locations,'' in \emph{Eur. Conf. Comput. Vis.}, 2020, pp. 649--665.

\bibitem{Box2Mask-C}
W.~Li, W.~Liu, J.~Zhu, M.~Cui, R.~Yu, X.~Hua, and L.~Zhang, ``Box2mask: Box-supervised instance segmentation via level-set evolution,'' \emph{IEEE Trans. Pattern Anal. Mach. Intell.}, vol.~46, no.~7, pp. 5157--5173, 2024.

\bibitem{Luo_et_al}
Y.~Luo, J.~Han, Z.~Liu, M.~Wang, and G.-S. Xia, ``An elliptic centerness for object instance segmentation in aerial images,'' \emph{J. Remote Sens.}, vol. 2022, 2022.

\bibitem{PointRend}
A.~Kirillov, Y.~Wu, K.~He, and R.~Girshick, ``Pointrend: Image segmentation as rendering,'' in \emph{IEEE Int. Conf. Comput. Vis. Pattern Recognit.}, June 2020.

\bibitem{Mask_Scoring_R-CNN}
Z.~Huang, L.~Huang, Y.~Gong, C.~Huang, and X.~Wang, ``Mask scoring r-cnn,'' in \emph{IEEE Int. Conf. Comput. Vis. Pattern Recognit.}, June 2019.

\bibitem{DCTC}
Z.~Chen, T.~Liu, X.~Xu, J.~Leng, and Z.~Chen, ``Dctc: Fast and accurate contour-based instance segmentation with dct encoding for high-resolution remote sensing images,'' \emph{IEEE J. Sel. Top. Appl. Earth Observ. Remote Sens.}, vol.~17, pp. 8697--8709, 2024.

\bibitem{SCNet}
T.~Vu, H.~Kang, and C.~D. Yoo, ``Scnet: Training inference sample consistency for instance segmentation,'' \emph{AAAI Conf. Artif. Intell.}, vol.~35, no.~3, pp. 2701--2709, May 2021.

\bibitem{Hybrid_Task_Cascade}
K.~Chen, J.~Pang, J.~Wang, Y.~Xiong, X.~Li, S.~Sun, W.~Feng, Z.~Liu, J.~Shi, W.~Ouyang, C.~C. Loy, and D.~Lin, ``Hybrid task cascade for instance segmentation,'' in \emph{IEEE Int. Conf. Comput. Vis. Pattern Recognit.}, June 2019.

\bibitem{Three_Module_ESWA}
W.~Ye, W.~Zhang, W.~Lei, W.~Zhang, X.~Chen, and Y.~Wang, ``Remote sensing image instance segmentation network with transformer and multi-scale feature representation,'' \emph{Expert Syst. Appl.}, vol. 234, p. 121007, 2023.

\bibitem{ARE-Net}
X.~Zeng, S.~Wei, J.~Shi, and X.~Zhang, ``A lightweight adaptive roi extraction network for precise aerial image instance segmentation,'' \emph{IEEE Trans. Instrum. Meas.}, vol.~70, pp. 1--17, 2021.

\bibitem{iSAID}
S.~Waqas~Zamir, A.~Arora, A.~Gupta, S.~Khan, G.~Sun, F.~Shahbaz~Khan, F.~Zhu, L.~Shao, G.-S. Xia, and X.~Bai, ``isaid: A large-scale dataset for instance segmentation in aerial images,'' in \emph{IEEE Int. Conf. Comput. Vis. Pattern Recognit. Workshops}, June 2019.

\bibitem{NWPU}
H.~Su, S.~Wei, S.~Liu, J.~Liang, C.~Wang, J.~Shi, and X.~Zhang, ``Hq-isnet: High-quality instance segmentation for remote sensing imagery,'' \emph{Remote Sens.}, vol.~12, no.~6, 2020.

\bibitem{SSDD}
T.~Zhang, X.~Zhang, J.~Li, X.~Xu, B.~Wang, X.~Zhan, Y.~Xu, X.~Ke, T.~Zeng, H.~Su, I.~Ahmad, D.~Pan, C.~Liu, Y.~Zhou, J.~Shi, and S.~Wei, ``Sar ship detection dataset (ssdd): Official release and comprehensive data analysis,'' \emph{Remote Sens.}, vol.~13, no.~18, 2021.

\bibitem{ResNet}
K.~He, X.~Zhang, S.~Ren, and J.~Sun, ``Deep residual learning for image recognition,'' in \emph{IEEE Int. Conf. Comput. Vis. Pattern Recognit.}, June 2016.

\bibitem{ViT}
A.~Dosovitskiy, L.~Beyer, A.~Kolesnikov, D.~Weissenborn, X.~Zhai, T.~Unterthiner, M.~Dehghani, M.~Minderer, G.~Heigold, S.~Gelly, J.~Uszkoreit, and N.~Houlsby, ``An image is worth 16x16 words: Transformers for image recognition at scale,'' \emph{Int. Conf. Learn. Represent.}, 2021.

\bibitem{YOLACT}
D.~Bolya, C.~Zhou, F.~Xiao, and Y.~J. Lee, ``Yolact: Real-time instance segmentation,'' in \emph{IEEE Int. Conf. Comput. Vis.}, October 2019.

\bibitem{Shi_and_Zhang}
F.~Shi and T.~Zhang, ``An anchor-free network with box refinement and saliency supplement for instance segmentation in remote sensing images,'' \emph{IEEE Geosci. Remote Sens. Lett.}, vol.~19, pp. 1--5, 2022.

\bibitem{FB-ISNet}
H.~Su, P.~Huang, J.~Yin, and X.~Zhang, ``Faster and better instance segmentation for large scene remote sensing imagery,'' in \emph{IEEE Int. Geosci. Remote Sens. Symp.}, 2022, pp. 2187--2190.

\bibitem{SA_R-CNN}
F.~Gao, Y.~Huo, J.~Wang, A.~Hussain, and H.~Zhou, ``Anchor-free sar ship instance segmentation with centroid-distance based loss,'' \emph{IEEE J. Sel. Top. Appl. Earth Observ. Remote Sens.}, vol.~14, pp. 11\,352--11\,371, 2021.

\bibitem{EMIN}
T.~Zhang and X.~Zhang, ``Enhanced mask interaction network for sar ship instance segmentation,'' in \emph{IEEE Int. Geosci. Remote Sens. Symp.}, 2022, pp. 3508--3511.

\bibitem{FL-CSE-ROIE}
------, ``A full-level context squeeze-and-excitation roi extractor for sar ship instance segmentation,'' \emph{IEEE Geosci. Remote Sens. Lett.}, vol.~19, pp. 1--5, 2022.

\bibitem{MAI-SE-Net}
------, ``A mask attention interaction and scale enhancement network for sar ship instance segmentation,'' \emph{IEEE Geosci. Remote Sens. Lett.}, vol.~19, pp. 1--5, 2022.

\bibitem{LFG-Net}
S.~Wei, X.~Zeng, H.~Zhang, Z.~Zhou, J.~Shi, and X.~Zhang, ``Lfg-net: Low-level feature guided network for precise ship instance segmentation in sar images,'' \emph{IEEE Trans. Geosci. Remote Sensing}, vol.~60, pp. 1--17, 2022.

\bibitem{YOLOv7}
C.-Y. Wang, A.~Bochkovskiy, and H.-Y.~M. Liao, ``Yolov7: Trainable bag-of-freebies sets new state-of-the-art for real-time object detectors,'' in \emph{IEEE Int. Conf. Comput. Vis. Pattern Recognit.}, June 2023, pp. 7464--7475.

\bibitem{S2ANet}
J.~Han, J.~Ding, J.~Li, and G.-S. Xia, ``Align deep features for oriented object detection,'' \emph{IEEE Trans. Geosci. Remote Sensing}, vol.~60, pp. 1--11, 2022.

\bibitem{Oriented_Reppoints}
W.~Li, Y.~Chen, K.~Hu, and J.~Zhu, ``Oriented reppoints for aerial object detection,'' in \emph{IEEE Int. Conf. Comput. Vis. Pattern Recognit.}, June 2022, pp. 1829--1838.

\end{thebibliography}


\begin{IEEEbiography}
    [{\includegraphics[width=1.0in,height=1.0in,clip,keepaspectratio]
    {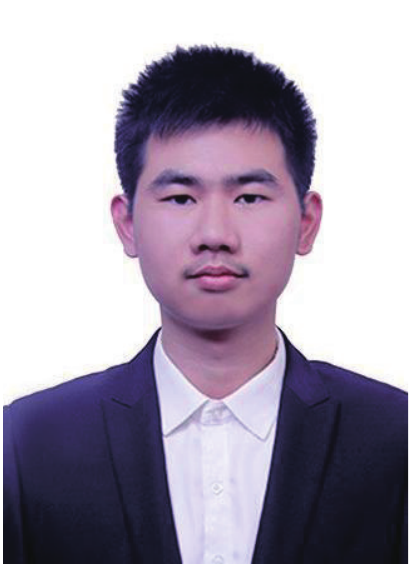}}]{Zhen Zhou}
    received the B.S. degree in automation from Hunan University, Changsha, in 2021. He is currently pursuing the Ph.D. degree at the State Key Laboratory of Multimodal Artificial Intelligence Systems, Institute of Automation, Chinese Academy of Sciences, Beijing. His research interests include computer vision and deep learning.
\end{IEEEbiography}


\begin{IEEEbiography}
    [{\includegraphics[width=1.0in,height=1.0in,clip,keepaspectratio]
    {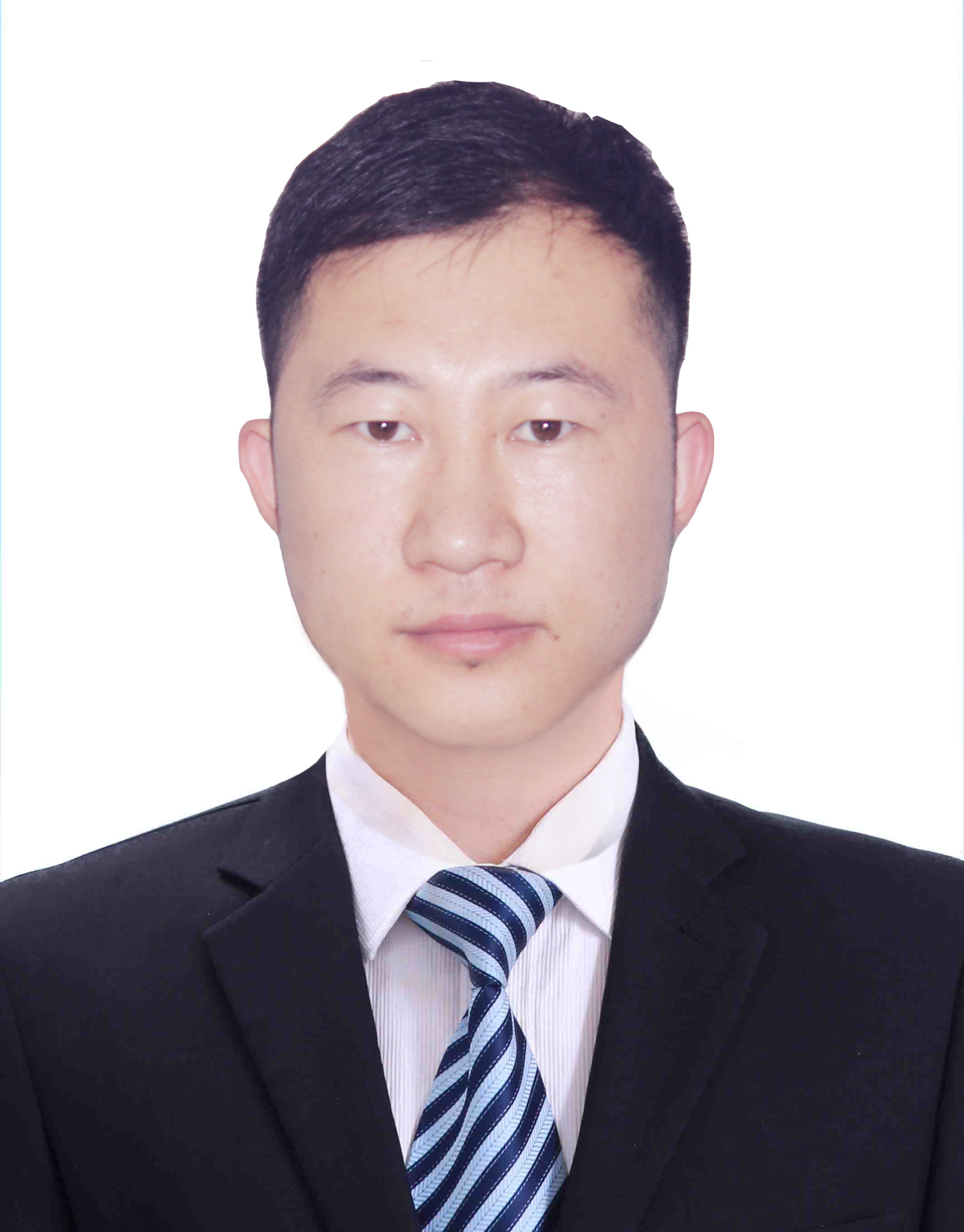}}]{Junfeng Fan}
    is currently an Associate Professor of the State Key Laboratory of Multimodal Artificial Intelligence Systems at the Institute of Automation, Chinese Academy of Sciences (IACAS), Beijing, China. His research interests include robot vision and industrial robotics.
\end{IEEEbiography}


\begin{IEEEbiography}
    [{\includegraphics[width=1.0in,height=1.0in,clip,keepaspectratio]
    {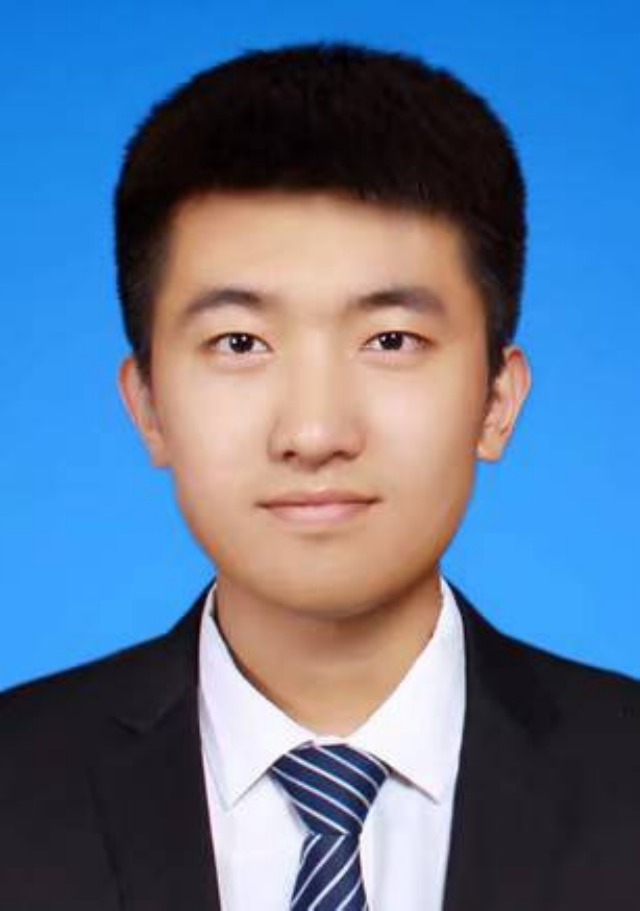}}]{Yunkai Ma}
    received the B.S. degree in intelligent science and technology from Qingdao University, Shandong Province, China, in 2017 and the Ph.D. degree in technology of computer applications from the Institute of Automation, Chinese Academy of Sciences, Beijing, in 2023. His research interests include computer vision and robotics.
\end{IEEEbiography}


\begin{IEEEbiography}
    [{\includegraphics[width=1.0in,height=1.0in,clip,keepaspectratio]
    {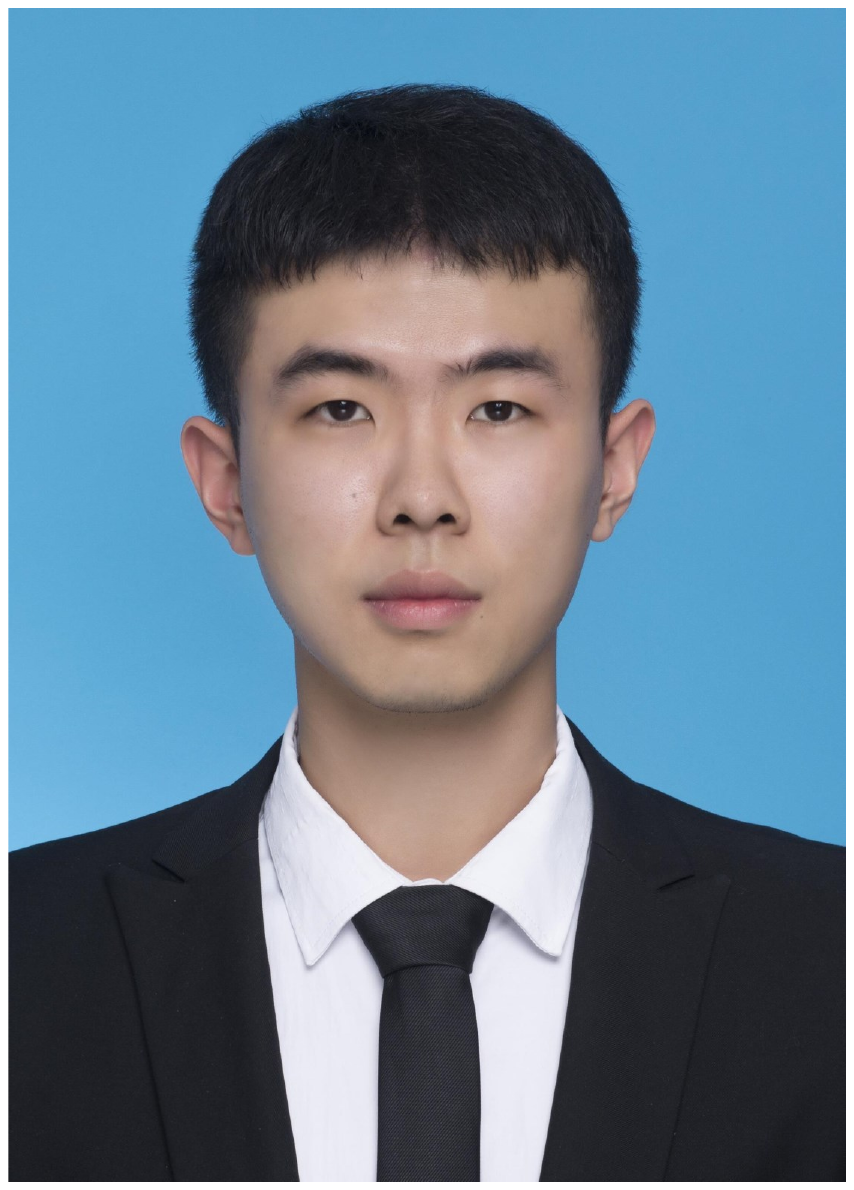}}]{Sihan Zhao}
    received the B.S. degree from Shandong University, Jinan, Shandong, China, in 2022. He is currently pursuing the Ph.D. degree in technology of computer applications with the School of Artificial Intelligence, University of Chinese Academy of Sciences, Beijing, China. His research interests include machine vision and
    robot control.
\end{IEEEbiography}


\begin{IEEEbiography}
    [{\includegraphics[width=1.0in,height=1.0in,clip,keepaspectratio]
    {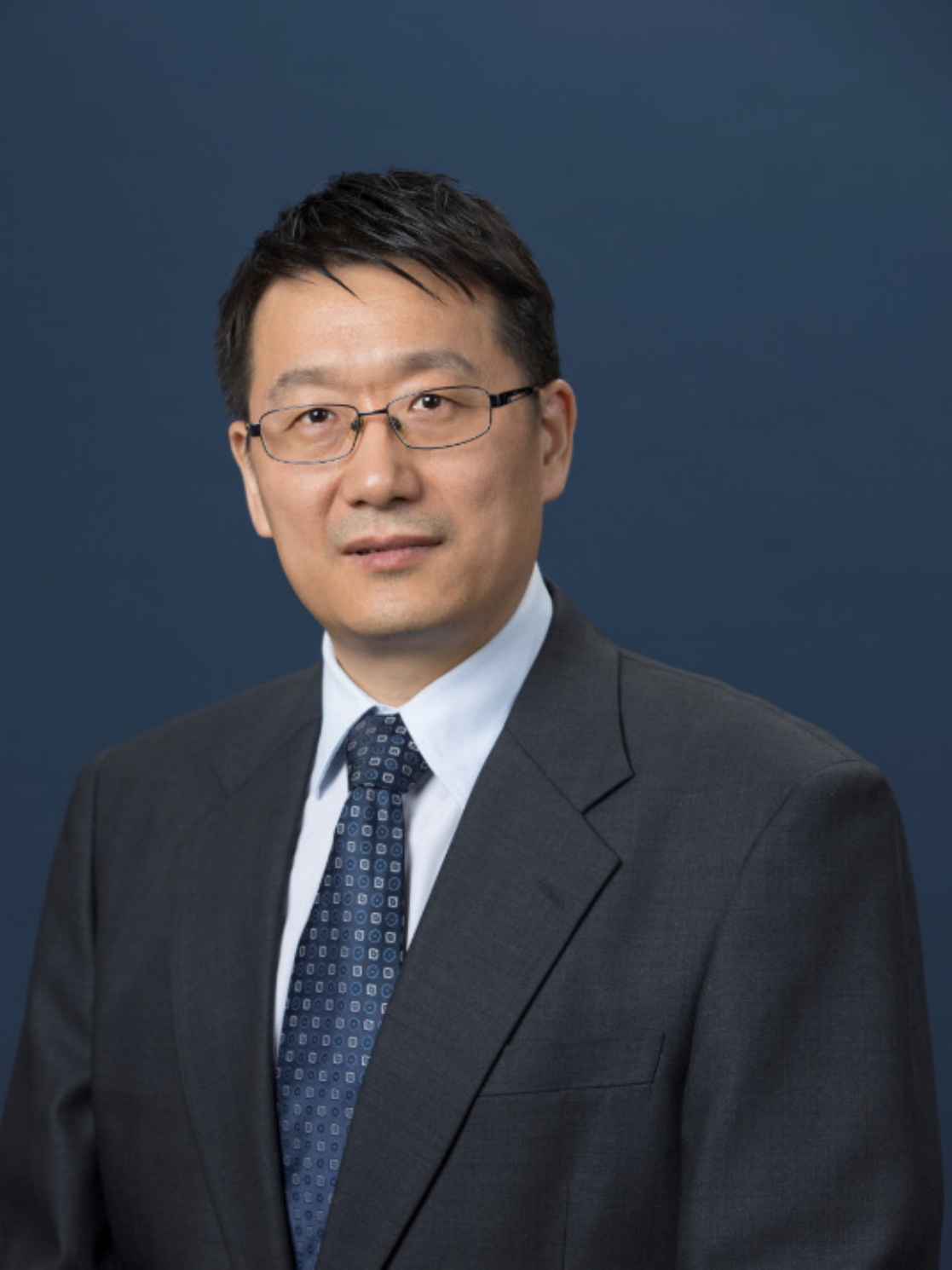}}]{Fengshui Jing}
    is currently a Professor in the State Key Laboratory of Multimodal Artificial Intelligence Systems, Institute of Automation, Chinese Academy of Sciences (IACAS), Beijing, China. His research interests include robotics, computer vision and manufacturing systems.
\end{IEEEbiography}


\begin{IEEEbiography}
    [{\includegraphics[width=1.0in,height=1.0in,clip,keepaspectratio]
    {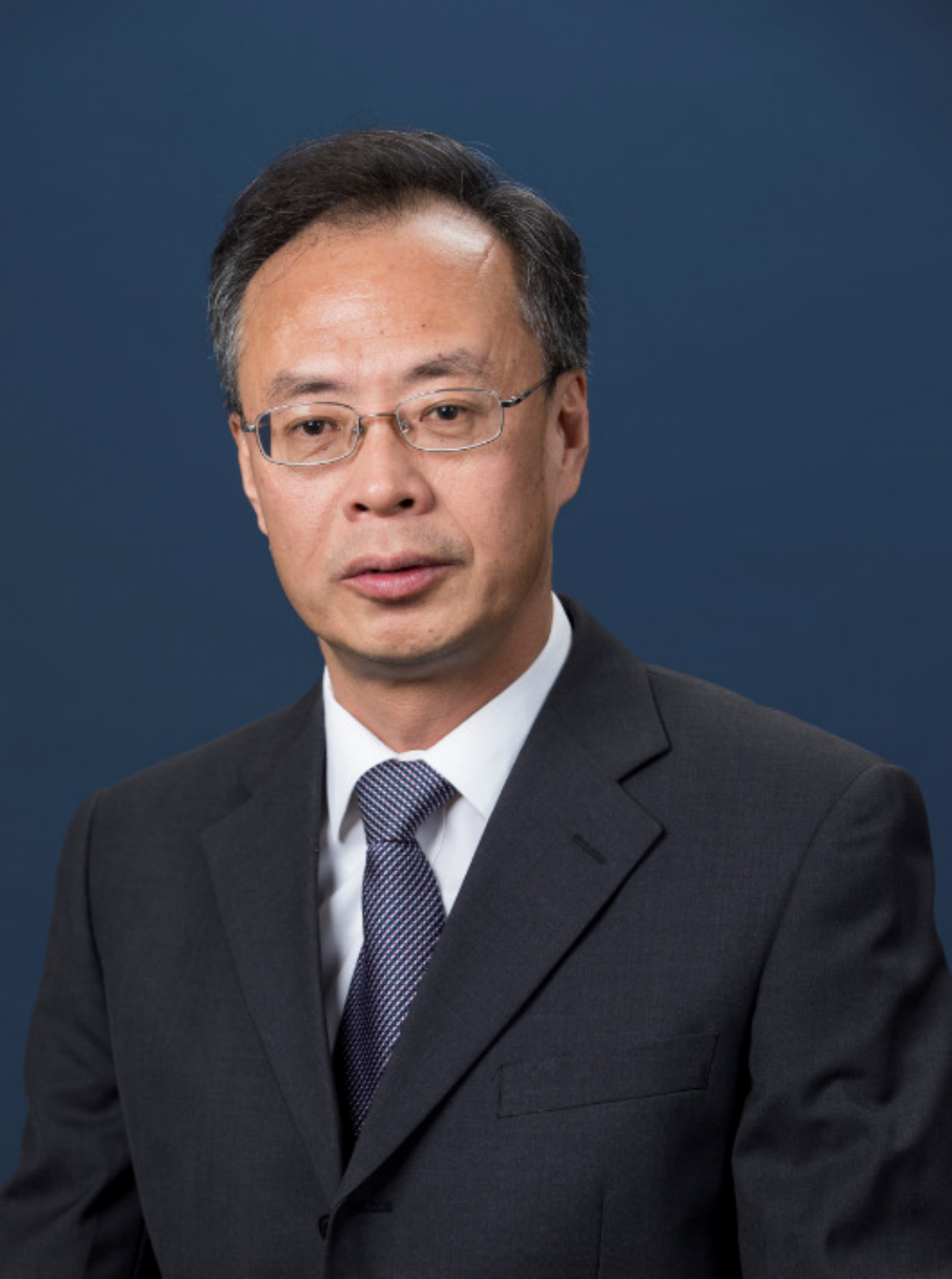}}]{Min Tan}
    is currently a Professor in the State Key Laboratory of Multimodal Artificial Intelligence Systems, Institute of Automation, Chinese Academy of Sciences (IACAS), Beijing, China. He has authored or coauthored more than 300 papers in journals, books, and conference proceedings. His research interests include robotics and intelligent systems.
\end{IEEEbiography}

\end{document}